\pdfoutput=1  
\documentclass{article} 
\usepackage{preprint,times}

\usepackage{amsmath,amsfonts,bm}

\def\eqref#1{equation~\ref{#1}}

\def\1{\bm{1}}

\DeclareMathAlphabet{\mathsfit}{\encodingdefault}{\sfdefault}{m}{sl}
\SetMathAlphabet{\mathsfit}{bold}{\encodingdefault}{\sfdefault}{bx}{n}

\usepackage{hyperref}
\usepackage{url}
\usepackage{booktabs}
\usepackage{tabularx}
\usepackage{graphicx}
\usepackage{parskip}
\usepackage{enumitem}
\usepackage{caption}
\newcommand{\numrows}{134}
\newcommand{\numckpts}{86}

\usepackage[labelfont=bf, margin=1.5em]{caption}

\title{Measuring the Inference Energy Efficiency \\of Open-Weight LLMs across Benchmarks, Reasoning Modes, and Hardware}

\author{Zeyu Yang \\
Department of Engineering Science \\
University of Oxford \\
\texttt{zeyu.yang@eng.ox.ac.uk}
\And
Wes Armour \\
Department of Engineering Science \\
University of Oxford \\
\texttt{wes.armour@oerc.ox.ac.uk}
}

\begin{document}

\maketitle

\begin{abstract}

Benchmark scores tell us what an LLM can do, but not how much energy is required to deliver that capability.
We directly measure this trade-off for 86 open-weight model checkpoints, evaluated on seven benchmarks and six hardware platforms ranging from a consumer GPU to an eight-GPU server. We find large differences in energy efficiency even among models with similar capability. Most configurations lie away from the score--energy Pareto frontier: they use a median of 2.6--4.8$\times$ more energy than another configuration that scores at least as well, with differences reaching 27$\times$. Reasoning can improve capability, but enabling thinking increases energy by about 3.7$\times$ on average, and the benefit depends strongly on the model family and output budget. Generated-token count is also an imperfect proxy for energy, misordering 18--24\% of model pairs. In contrast, FP8 reduces energy by roughly a quarter with little change in score, while changing hardware largely rescales energy without changing the relative ranking of models. These results show that capability alone gives an incomplete picture of efficiency. We argue that benchmark scores should be reported together with measured energy consumption under a specified serving regime and output budget.
\end{abstract}

\section{Introduction}

Two LLMs with similar benchmark scores can require very different amounts of energy to serve. Among the open-weight models we evaluate that score above 80\% on AIME, the GPU energy required per correct answer spans roughly \(33\times\), from 1.5 to 50 Wh depending on the model and the hardware serving it (Fig.~\ref{fig:energy_per_correct}; Appendix~\ref{app:per_correct}). Conventional leaderboards capture the difference in capability, but not the energy required to deliver it.

This distinction is becoming increasingly important as inference accounts for a growing share of AI computation. Around three fifths of the energy Google reports for machine learning is attributed to inference rather than training \citep{patterson2022plateau}, while the output lengths of reasoning models have grown rapidly \citep{emberson2025epochlength}. More broadly, global data-centre electricity consumption was approximately 415\,TWh in 2024 and is projected to more than double by 2030, with AI as a major driver \citep{iea2025energyai}. Evaluating model capability without considering the energy required to deliver it therefore gives an incomplete picture of inference efficiency.

Measuring this trade-off is not as simple as counting parameters, FLOPs, or generated tokens. These quantities are often used as proxies for inference cost, but do not uniquely determine physical energy consumption \citep{henderson2020systematic,dehghani2022efficiency,fernandez2025energy}. Direct measurements are also sensitive to serving conditions: batching, sequence length, hardware, precision, and the serving stack can all substantially change the energy required for the same workload \citep{saad2025intelligence,chung2026joules,pizzini2026sweetspot}. Meanwhile, capability leaderboards typically report benchmark scores without the corresponding measured energy consumption. The capability--energy trade-off across a broad set of current LLMs under a common high-utilisation serving regime therefore remains poorly characterised.

In this work, we measure this trade-off directly. We evaluate \numckpts{} open-weight checkpoints, corresponding to \numrows{} serving configurations, on seven scored benchmarks spanning instruction following, knowledge, science, mathematics, coding, and tool use. Models are served with saturated continuous batching on six hardware platforms ranging from a consumer GPU to an eight-GPU server. Rather than assigning each model a fixed energy-per-token value, we meter GPU energy consumption using each benchmark's own traffic, including the additional tokens generated by reasoning modes. We then compare benchmark score against measured energy consumption and analyse the resulting score--energy Pareto fronts. 

\begin{figure}[ht]
  \centering
  \includegraphics[width=\linewidth]{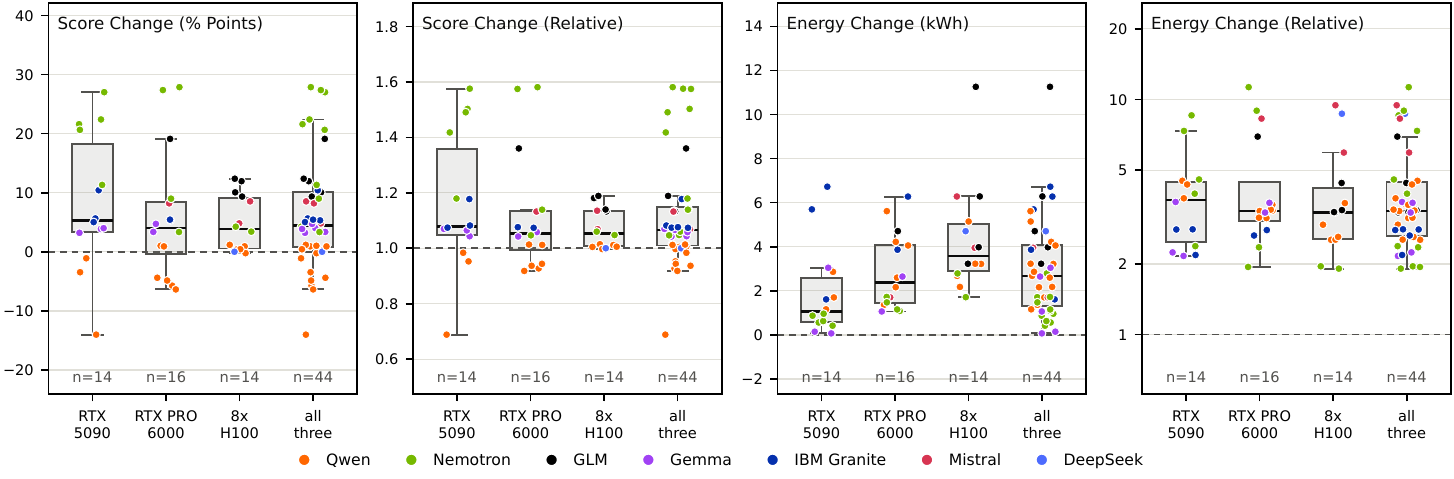}
  \vskip -0.1in
  \caption{\textbf{Effect of enabling reasoning with the same weights on the same hardware.} Each point compares one hybrid checkpoint with thinking disabled and enabled on its tier's reference platform (44 pairs), coloured by lineage; boxes show quartiles by platform and pooled. \emph{Left:} change in suite score, shown in points and as a ratio. \emph{Right:} change in GPU energy for one pass over the suite, shown in kWh and as a ratio (log scale). }
  \vskip -0.1in
  \label{fig:thinking_pairs}
\end{figure}

\begin{figure}[ht]
  \centering
  \vskip -0.1in
  \includegraphics[width=\linewidth]{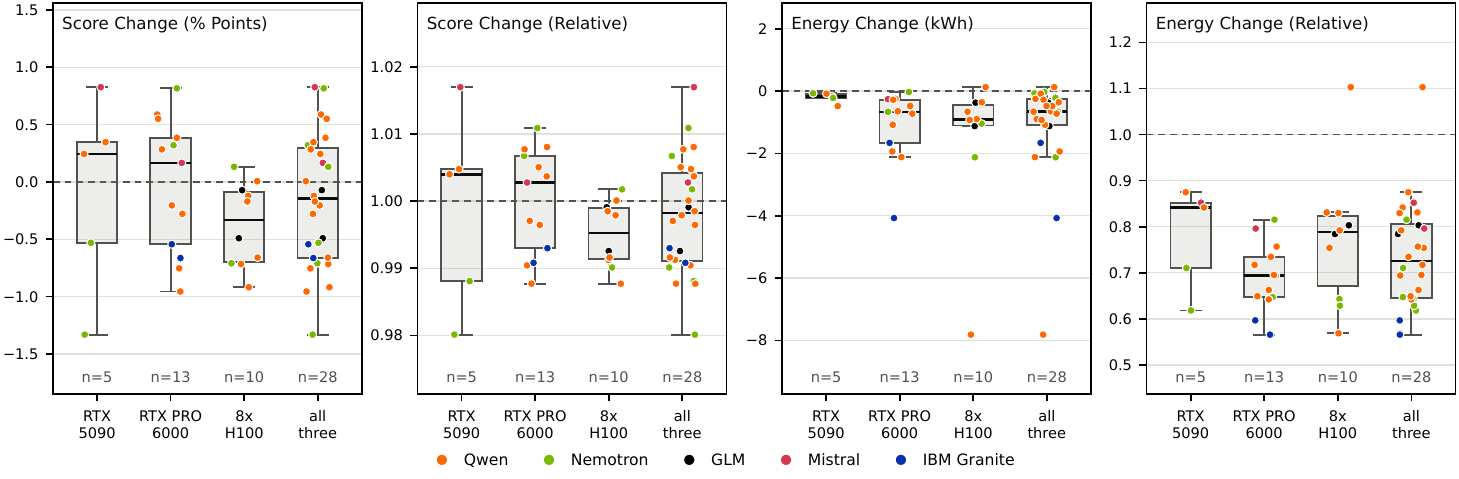}
  \vskip -0.1in
  \caption{\textbf{Effect of replacing BF16 weights with vendor-released FP8 weights on the same hardware.}
  Each point compares the same base model and reasoning mode in vendor-released BF16 and FP8 variants on its tier's reference platform (28 matched pairs), coloured by lineage; boxes show quartiles by platform and pooled. \emph{Left:} change in suite score, in points and as a ratio. \emph{Right:} change in GPU energy consumption, in kWh and as a ratio.}
  \vskip -0.05in
  \label{fig:quant_twins}
\end{figure}

\paragraph{Contributions}
Our main contributions and findings are:

\begin{itemize}[itemsep=1pt, topsep=0pt, parsep=0pt]
    \item \textbf{A large-scale measurement of capability against directly measured inference energy.} 
    In the main cross-model comparison, only 21 of 82 configurations lie on a Pareto front; off-front configurations use a median of \(2.6\)--\(4.8\times\) the energy of an alternative scoring at least as well, and up to \(27\times\) (Fig.~\ref{fig:frontier_row}).

    \item \textbf{A controlled characterisation of the energy cost of reasoning and longer outputs.}
    Across 44 paired hybrid configurations, enabling thinking increases energy by about \(3.7\times\), while the capability gain depends strongly on the model lineage (Fig.~\ref{fig:thinking_pairs}). Increasing the output budget can recover substantial additional capability, but at up to \(6.1\times\) the energy; late tokens are also \(2.0\)--\(4.9\times\) more energy-intensive (Fig.~\ref{fig:overthink}).

    \item \textbf{Evidence that commonly used inference-cost proxies can give a different picture from physical energy.}
    Generated-token count misorders energy for 18--24\% of model pairs, and score--token Pareto fronts retain only 10 of the 25 models selected by the corresponding score--energy fronts (Fig.~\ref{fig:verbosity}). Parameter count is an even weaker predictor of energy.

    \item \textbf{A cross-platform analysis of deployment choices.}
    Vendor-provided FP8 weights reduce GPU energy by roughly a quarter with no measurable change in aggregate score (Fig.~\ref{fig:quant_twins}), while the same configurations consume \(0.5\)--\(2.2\times\) as much energy across hardware platforms but preserve almost the same energy ranking (\(\rho = 0.99\); Fig.~\ref{fig:xplat_96g}).
\end{itemize}

\section{Preliminaries: What Serving Energy Depends On}

Serving a language model has two main phases. Prefill processes the input sequence in parallel and is typically compute-intensive, while decode generates tokens autoregressively and is often memory-bandwidth-bound at small batch sizes \citep{pope2023efficiently,patel2024splitwise}. Each decode step accesses the model parameters and accumulated KV state, making memory movement a major cost. Continuous batching improves utilisation by serving many active requests together and replacing completed requests as execution proceeds \citep{yu2022orca,kwon2023efficient}. This allows model execution to be shared across more requests and increases throughput. Since GPU power typically grows much less than throughput as concurrency increases \citep{patel2024polca}, higher utilisation generally reduces energy per token until KV-cache capacity limits further batching \citep{kwon2023efficient,chung2026joules}.

Energy per token is therefore not a property of the model alone. It depends on the model, hardware, serving system, and workload. Batch size can change per-token energy by several-fold \citep{chung2026joules}, while input and output lengths can produce much larger variation on the same model and GPU \citep{pizzini2026sweetspot}. Long outputs are especially costly because they generate more tokens while increasing KV-cache usage and reducing achievable concurrency \citep{chung2026joules}. In finite workloads, concurrency can also fall as shorter requests finish, so late tokens may be generated in smaller batches even when the GPU remains highly powered. A single energy-per-token value therefore cannot reliably represent the cost of a benchmark, and simple models based only on prompt and output token counts can incur substantial error under batched serving \citep{pizzini2026sweetspot,dutt2025fine}.

Meaningful comparisons therefore require energy to be measured under a defined workload, serving regime, and hardware platform. This motivates our use of direct GPU level energy measurements under high-utilisation continuous batching, rather than assigning each model a fixed energy-per-token value. Our measurement boundary excludes host and data-centre overhead. For context, Google's production measurements report total serving energy at roughly \(1.7\times\) active accelerator energy after accounting for host systems, idle capacity, and infrastructure overhead \citep{elsworth2025google}. Section~\ref{sec:methodology} describes the workload construction and measurement procedure.

\section{Related Work}

\paragraph{Energy benchmarking}
Several benchmarks directly measure LLM inference energy. ML.ENERGY evaluates more than 40 models and workloads \citep{chung2025mlenergy,chung2026joules}, while MLPerf Power reports system efficiency subject to workload-specific accuracy requirements \citep{tschand2024mlperfpower}. The AI Energy Score measures energy per query at batch size~1 \citep{aienergyscore2025,luccioni2024powerhungry}. These efforts establish the value of direct measurement, but do not jointly characterise capability and energy across a broad model set under a common high-utilisation serving regime, where batching, sequence shape, software, and hardware can substantially affect energy \citep{fernandez2025energy,chung2026joules,saad2025intelligence}. 

\paragraph{Serving efficiency and reasoning length}
Systems work reduces inference energy through frequency, parallelism, batching, and scheduling \citep{you2023zeus,patel2024polca,stojkovic2025dynamollm}, while other work controls or shortens reasoning traces \citep{snell2024scaling,muennighoff2025s1,aggarwal2025l1,chen2024overthinking}. Response-length studies further show substantial variation across models and increasingly long reasoning traces \citep{poddar2025brevity,emberson2025epochlength}. Our focus is complementary: we measure how these choices affect the energy required to deliver benchmark capability. 

\paragraph{Energy proxies and metrics}
Several works connect capability to estimated inference cost. Token Arena estimates joules per correct answer from hardware power assumptions \citep{gao2026tokenarena}; \citet{oviedo2026energyuse} estimate per-query energy from published throughput and power data; and Capability Frontier and OckBench use monetary cost or token counts \citep{fowler2026capabilityfrontier,du2025ockbench}. Analytical models similarly estimate energy from FLOPs or token counts \citep{faiz2024llmcarbon,wilkins2024offline,pizzini2026sweetspot}, although accuracy can degrade as serving conditions vary \citep{pizzini2026sweetspot,dutt2025fine,fu2025llmco2}. Prior work therefore cautions against universal efficiency proxies \citep{schwartz2020greenai,henderson2020systematic,dehghani2022efficiency}. We instead retain capability and measured energy as separate objectives and analyse their Pareto frontier.

\paragraph{Joint energy and capability measurement}
A smaller body of work directly measures energy while evaluating model quality. Intelligence per Watt evaluates more than 20 models across eight accelerators, primarily at batch size~1 \citep{saad2025intelligence}; as Appendix~\ref{app:batch1} shows, this regime can have substantially higher energy per token than saturated serving. \citet{jin2025energycostreasoning} study one model family at fixed batch size, while \citet{stuhlmann2025bench360,ali2026sustainability} consider smaller model sets under light-load conditions. \citet{caravaca2025prompts} combine published capability scores with measured energy for a synthetic workload. In vision, \citet{yang2025hidden} jointly measured accuracy and energy for 1,200 architectures at saturating batch sizes.

Together, these leave open the capability--energy frontier across current LLMs under high-utilisation serving. We address this with scored benchmark traffic, saturated continuous batching, and controlled comparisons of reasoning mode, output budget, precision, and hardware.

\section{Methodology}
\label{sec:methodology}

\paragraph{Models}
We evaluate \numckpts{} open-weight checkpoints from eleven model families: Qwen, DeepSeek, Llama, Gemma, GLM, IBM Granite, Mistral, Phi, Nemotron, MiniMax, and gpt-oss. The checkpoints were released between July 2024 and August 2026, span 0.8B--400B parameters, and include 43 dense and 43 mixture-of-experts models. Within each family, we include instruction-tuned checkpoints across the available size range, subject to compatibility with our pinned serving stack. Vendor-published FP8 and NVFP4 weight variants are included as separate checkpoint entries.

Our unit of study is a \emph{serving configuration}, rather than the model weights alone. For hybrid models, reasoning mode is controlled through the chat template and changes both capability and generation length, so each mode forms a separate configuration. Single-mode checkpoints contribute one configuration, giving \numrows{} in total.

Sampling follows the generation configuration shipped with each checkpoint or the settings recommended in its model card; where neither is available, we use temperature 0.6 and top-$p$ 0.95. This preserves the intended serving behaviour of each checkpoint rather than standardising decoding across models. The full model list and sampling provenance are given in Appendix~\ref{app:models}.

\paragraph{Hardware}
Models are grouped by weight footprint into three memory tiers: 32, 96, and 512\,GB: the weight budget available to the model while retaining headroom for the KV cache. Each tier is assigned only to platforms with sufficient memory (Table~\ref{tab:platforms}). We examine CPU weight offloading separately in Appendix~\ref{app:offload}. The GB10 is capped at 96 of its 128\,GB unified memory, leaving the remainder for the operating system and serving stack.

For the main cross-model comparison, the 32, 96, and 512\,GB tiers use the RTX~5090, RTX PRO~6000, and 8$\times$H100 as their reference platforms, respectively. The 96\,GiB model set is additionally served on three other platforms, spanning three GPU generations and three memory technologies, enabling like-for-like comparison of the same serving configurations across hardware. Full system and software details are given in Appendix~\ref{app:hardware}.

\begin{table}[h]
\centering
\caption{Serving platforms. Multi-GPU platforms are NVLink-connected and use tensor parallelism, plus expert parallelism for MoE models on the 8-GPU node.}
\label{tab:platforms}
\small
\begin{tabular}{llllc}
\toprule
Platform & Class & Arch. & Memory  & Model Tier \\
\midrule
GeForce RTX 5090        & desktop     & Blackwell & 32\,GB GDDR7            & 32  \\
2$\times$ RTX A6000     & workstation & Ampere    & 2$\times$48\,GB GDDR6   & 96  \\
GH200                   & server      & Hopper    & 96\,GB HBM3             & 96  \\
GB10 (DGX Spark)        & desktop     & Blackwell & 128\,GB unified LPDDR5X & 96  \\
RTX PRO 6000 Blackwell  & workstation & Blackwell & 96\,GB GDDR7            & 96  \\
8$\times$ H100 SXM      & server      & Hopper    & 8$\times$80\,GB HBM3    & 512 \\
\bottomrule
\end{tabular}
\end{table}

\paragraph{Benchmarks}
The seven-benchmark suite is summarised in Table~\ref{tab:suite}. All headline metrics use deterministic scoring without an LLM judge. MMLU-Pro and BFCL use stratified subsets sampled once with a fixed seed and then frozen as ID lists, so every configuration receives the same questions. We use multiple seeded attempts for the higher-variance GPQA and AIME benchmarks. AIME 2026 and LiveCodeBench provide recent evaluation items with lower exposure to older training corpora.

Benchmarks define the problems and scoring rules, but not a common generation budget. We therefore use benchmark-specific output caps held fixed across models, ranging from 4K to 32K tokens. A generation that reaches the cap is scored normally; if truncation removes the extractable answer, it is counted as incorrect. We do not resample truncated answers or assign partial credit. Instead, we report truncation rates, treating completion within the common budget as part of model performance. We separately examine the sensitivity of score and energy to larger output budgets in Fig.~\ref{fig:overthink}.

\begin{table}[h]
\centering
\caption{The benchmark suite. $k$ = seeded attempts. cap = maximum new tokens. $^{\dagger}$per turn.}
\label{tab:suite}
\small
\setlength{\tabcolsep}{5.9pt} 
\begin{tabular}{llrcrl}
\toprule
Benchmark & Ability & Items & $k$ & Cap & Headline metric \\
\midrule
IFEval~\citep{zhou2023ifeval}                  & instr. following      & 541     & 1 & 8K  & prompt-level strict acc. \\
MMLU-Pro (1k)~\citep{wang2024mmlupro}          & knowledge             & 1{,}000 & 1 & 16K & exact match \\
GPQA Diamond~\citep{rein2023gpqa}              & science               & 198     & 2 & 32K & exact match \\
MATH-500~\citep{lightman2024math500}           & math                  & 500     & 1 & 16K & exact match \\
AIME 25/26~\citep{aime}                        & competition math      & 60      & 4 & 32K & exact match \\
LiveCodeBench~\citep{jain2025livecodebench}    & coding                & 175     & 1 & 32K & pass@1 (executed) \\
BFCL v3~\citep{patil2025bfcl}                  & tool use              & 2{,}073 & 1 & 4K$^{\dagger}$ & macro avg. 17 categories \\
\bottomrule
\end{tabular}
\end{table}

\paragraph{Evaluation harness}
All benchmark harnesses communicate with the serving backend through a common OpenAI-compatible logging proxy. The proxy applies sampling parameters, output caps, and per-attempt seeds, and records prompt, cached, and generated token counts. Before evaluation, a smoke test checks the chat template, reasoning parser, answer extraction, and truncation behaviour. 

\paragraph{Deployment}
We serve all models with a pinned version of vLLM~\citep{kwon2023efficient} under saturated continuous batching. Continuous batching, chunked prefill, and prefix caching are enabled. We standardise the serving regime rather than a numerical concurrency: each platform retains the engine's default admission limit, while client concurrency is set to \(1.5\times\) that limit to maintain a continuous backlog and leave batching decisions to the server scheduler. Prefix-cache state is reset between benchmark runs, while reuse within a run is retained as part of the workload.

\paragraph{Energy measurement}
We use accelerator-level energy as our primary measurement boundary. On discrete GPUs, board power is sampled from NVIDIA's onboard sensors and integrated over time; multi-GPU energy is summed across devices \citep{yang2024accurate}. On the GH200 we use the GPU+HBM reading, while on the GB10 we use the GPU-rail energy counter. Package-level measurements for both superchips are recorded separately and reported in Appendix~\ref{app:energy}. CPU and host memory on the discrete-GPU systems, power-supply losses, cooling, and other infrastructure overheads are outside the primary measurement boundary.

Energy is measured from each benchmark's own request traffic rather than from a synthetic workload. We identify the active serving interval from server-side token counters and exclude harness startup and CPU-side scoring. The first 10\% and final 15\% of the active interval are removed to exclude ramp-up and drain-out, leaving the saturated steady-state region. 

\paragraph{Scoring and aggregation}
Each benchmark is scored using its standard metric (Table~\ref{tab:suite}) after any reasoning trace is removed by the model-specific parser. Scores are averaged over the \(k\) seeded attempts. Generated-token counts include reasoning tokens, so the additional generation produced by reasoning modes remains part of their measured cost.

We treat capability and energy as separate objectives rather than combining them into a score-per-energy ratio, which can favour inexpensive configurations with poor absolute capability. A configuration is Pareto efficient if no other configuration in the comparison achieves at least its score while using no more energy, with a strict improvement in at least one objective.

For the cross-benchmark summary, capability is the arithmetic mean of the seven headline scores, while energy is the geometric mean of energy per attempt across the seven benchmark components. The geometric mean gives each benchmark equal weight on a multiplicative scale rather than allowing the highest-energy workloads to dominate. Appendix~\ref{app:aggregation} analyses the effect of this aggregation.

\section{The Capability--Energy Frontier}
\label{sec:frontier}

\begin{figure}[ht]
  \centering
  \vskip -0.1in
  \includegraphics[width=\linewidth]{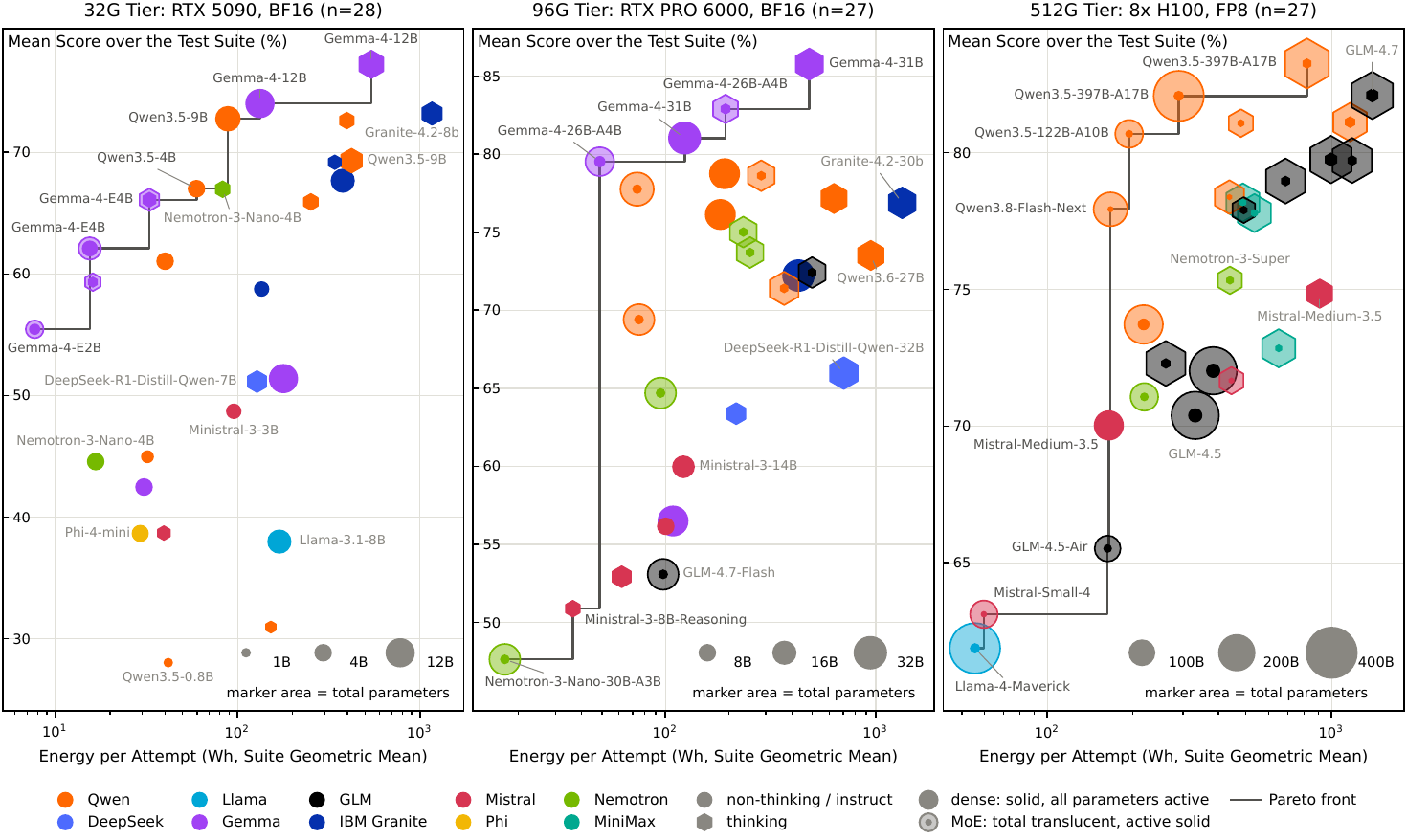}
  \vskip -0.1in

  \caption{\textbf{Capability--energy frontiers across the three memory tiers.} Each panel shows one tier on its reference platform (BF16 for 32/96\,GiB; FP8 for 512\,GiB). $x$: geometric-mean GPU energy per attempt; $y$: mean suite score. Marker encoding follows the legend; steps show the Pareto fronts. Fig.~\ref{fig:frontier_stack} gives the fully labelled version.}
  
  \vskip -0.05in
  \label{fig:frontier_row}
\end{figure}

We first compare capability with energy across the three reference tiers. Fig.~\ref{fig:frontier_row} places each serving configuration in the score--energy plane, where higher scores and lower energy are preferable.

\paragraph{Most configurations are dominated}
Across the three reference-tier comparisons, only 21 of 82 configurations lie on a Pareto frontier. For the remaining configurations, the cheapest alternative that scores at least as high uses a median of \(2.6\)--\(4.8\times\) less energy, depending on the tier; the largest difference is \(27\times\). On the RTX PRO~6000, for example, Granite-4.2-30b with thinking scores 77.6 at 1{,}307\,Wh, while Gemma-4-26B-A4B without thinking scores 79.5 at just 49\,Wh. Large differences in energy therefore remain even after capability is held approximately fixed.

\paragraph{The highest scores come at a steep energy premium}
Thinking configurations achieve the highest score on all three frontiers, but much of the final gain in capability is expensive. In every tier, a non-thinking configuration comes within five points of the highest score while using only a quarter to a sixth of its energy (Qwen3.5-9B, Gemma-4-31B, and Qwen3.5-122B-A10B). Overall, non-thinking configurations account for 15 of the 21 frontier members, compared with 6 of the 42 thinking configurations. We examine the effect of reasoning mode and the output budget in Section~\ref{sec:reasoning_budget}.

\begin{figure}[ht]
  \centering
  \vskip -0.05in
  \includegraphics[width=\linewidth]{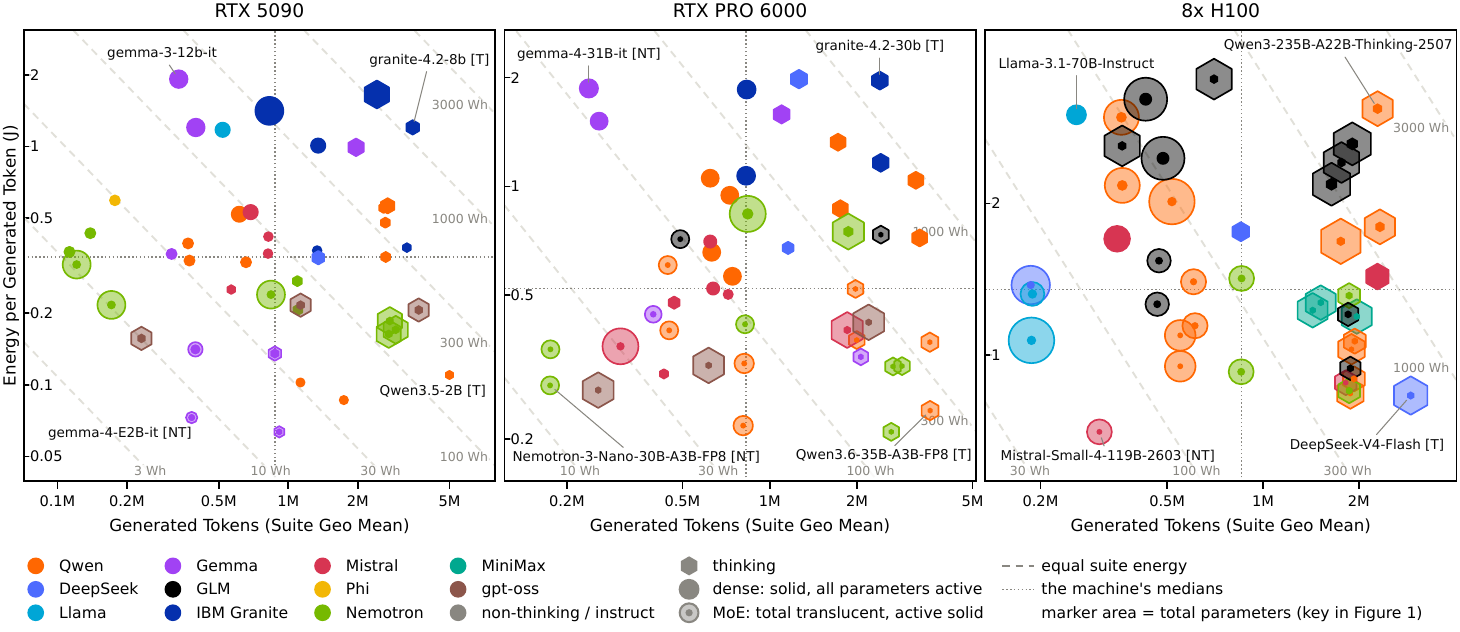}
  \vskip -0.1in

  \caption{\textbf{Energy varies through both generation length and energy per token.} Each panel shows all complete configurations on a reference platform, across weight formats. $x$: geometric-mean generated tokens per attempt; $y$: GPU energy per generated token, including prefill. Their product gives the suite energy in Fig.~\ref{fig:frontier_row}. Dashed diagonals show equal energy; dotted lines show platform medians. Marker encoding follows Fig.~\ref{fig:frontier_row}.}
  \vskip -0.05in
  \label{fig:verbosity}
\end{figure}

Energy correlates more strongly with generated-token count (Spearman \(\rho = 0.70\)--\(0.79\)) than with active parameter count (\(\rho = 0.28\)--\(0.52\)), but token count alone still does not explain which configurations are energy efficient. Fig.~\ref{fig:verbosity} separates these two components.

\paragraph{Generated tokens are not a reliable proxy for energy}
Across the three reference platforms, generated-token counts vary by \(6\)--\(14\times\) between the 90th and 10th percentiles, while energy per generated token varies independently by \(3\)--\(10\times\). The two quantities are nearly uncorrelated (Pearson \(r=-0.23\) to \(+0.02\)), so generating more tokens does not imply that each token is more expensive.

Token count nevertheless captures part of the variation in total energy, explaining 60--85\% of its log-scale variance. Much of this association reflects the separation between thinking and non-thinking configurations: within either group, variation in token count and energy per token are of comparable importance. As a ranking proxy, however, generated tokens misorder energy for 18--24\% of configuration pairs. Two non-thinking configurations with nearly identical token counts---Gemma-3-12B and Gemma-4-E2B on the RTX~5090---differ by \(23\times\) in energy.

This difference is large enough to change which models appear efficient. On the three tiers, the score--token frontier shares only 4 of 11, 3 of 6, and 3 of 8 members with the corresponding score--energy frontier. It excludes configurations that generate many tokens cheaply, such as Gemma-4-26B-A4B and Qwen3.5-122B-A10B, while admitting Qwen3-235B-A22B-Instruct, which generates relatively few tokens at roughly 3\,J each. Token-based cost measures \citep{du2025ockbench} can therefore identify a substantially different set of efficient models from direct energy measurement.

\section{Reasoning and Output Budget}
\label{sec:reasoning_budget}
The frontier analysis suggests that reasoning can reach higher capability, but often at substantially higher energy. We isolate this effect using hybrid checkpoints that can be served with and without reasoning while keeping the weights and hardware fixed.

\paragraph{Reasoning is consistently expensive, but its capability gain is not}
Across the 44 paired configurations, enabling thinking increases the suite score by a median of 4.5 points (mean 6.3; 95\% CI 3.6--9.1), while multiplying the energy of a suite pass by \(3.7\times\) on average. Eight of the 44 pairs lose score despite consuming more energy.

The score gain varies much more by model lineage than by hardware. We find no detectable platform effect on either the score or energy change (Kruskal--Wallis \(p=0.60\) and \(0.88\), respectively), whereas the difference between lineages is significant (\(p<0.001\)). Nemotron hybrids gain a median of 20.6 points and none loses score; GLM hybrids gain 12.0 points in the median. Qwen behaves very differently: its median change is \(-0.7\) points, with 8 of 14 pairs scoring lower when thinking is enabled. More generally, the gain from reasoning decreases as the corresponding non-thinking score rises (Spearman \(\rho=-0.69\)).

These losses, however, are entangled with the output budget. Enabling thinking raises the average share of requests truncated at the benchmark cap from 3\% to 17\%. For the four Qwen configurations with the largest score losses, 25--62\% of requests are truncated. Under the optimistic clean-completion bound of Fig.~\ref{fig:frontier_stack_clean}, these configurations would instead reach suite scores of 78--88. The measured benefit of reasoning is therefore conditional not only on the model lineage, but also on how much generation budget it is given.

\begin{figure}[ht]
  \centering
  \vskip -0.1in
  \includegraphics[width=\linewidth]{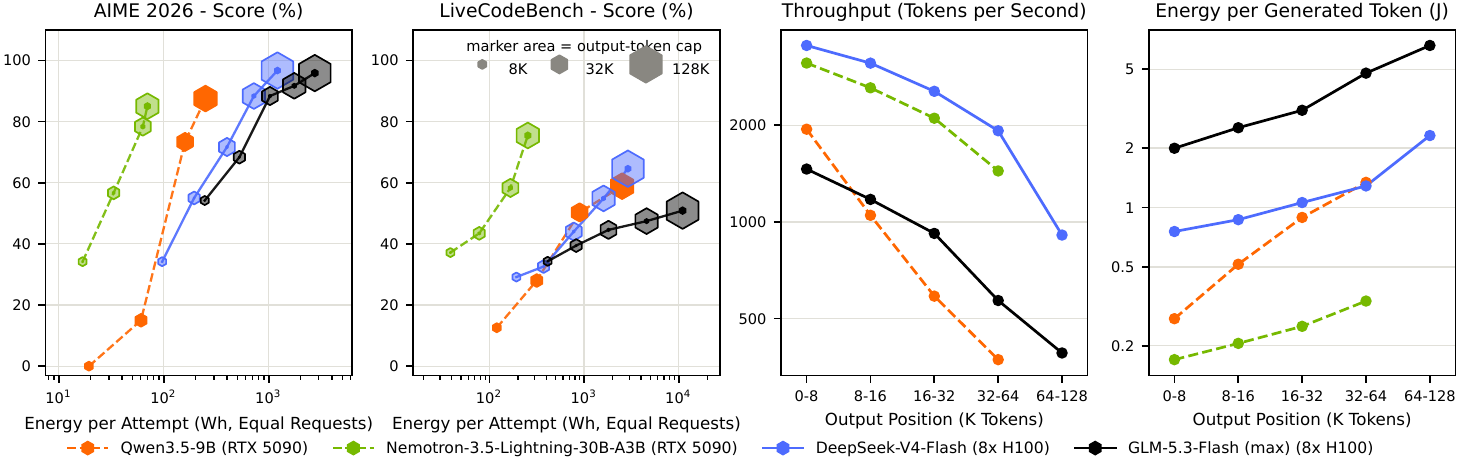}
  \vskip -0.1in

  \caption{\textbf{Increasing the output budget can recover capability at higher energy cost.} Four thinking configurations are rerun from 8K to 128K output tokens; the main evaluation uses 32K. \emph{Left:} score against GPU energy per attempt, with marker area proportional to the output cap; energy is normalised to the request count. \emph{Right:} throughput and energy per generated token by output position. Dashed lines: RTX~5090; solid lines: 8$\times$H100.}

  \vskip -0.05in
  \label{fig:overthink}
\end{figure}

\paragraph{The output budget can change both scores and model rankings}
Fig.~\ref{fig:overthink} reruns four thinking configurations beyond the 32K cap used in the main evaluation. Increasing the cap beyond 32K improves AIME 2026 by 7--25 points and LiveCodeBench by 6--21 points, but requires \(1.1\)--\(6.1\times\) as much energy. The return on this additional energy varies sharply between models. For the final budget doubling, Nemotron-3.5-Lightning requires 1\,Wh for each additional AIME point, compared with 239\,Wh for GLM-5.3-Flash; on LiveCodeBench the values are 5 and 1{,}869\,Wh per point.

More importantly, the budget can change the relative ordering of models. DeepSeek-V4-Flash, for example, ranks last among the four configurations on AIME 2026 at 32K but first at the largest output cap. The scores in the main frontier should therefore be interpreted as capability \emph{under a common output budget}, rather than as estimates of each model's unconstrained maximum capability.

\paragraph{Late tokens become more expensive as the KV state grows}
Within the same runs, energy per generated token rises \(2.0\)--\(4.9\times\) from the first 8K window of output to the last, while throughput falls to \(0.19\)--\(0.46\times\). Longer sequences can increase cost in two ways. Their growing inference state increases memory pressure and reduces achievable concurrency; in attention layers whose per-token work grows with context length, later tokens can also require more state access. In our serving logs, KV-cache usage remains close to capacity while the number of running sequences falls to \(0.19\)--\(0.37\times\); GPU power, however, remains at \(0.78\)--\(0.94\times\) of its initial level. Late tokens therefore require more work per sequence while being generated at lower concurrency, making them substantially more energy-intensive than early tokens.

\section{Deployment Choices}
\label{sec:deployment}

The previous sections compare different models and reasoning behaviours. We next ask how much energy can be changed through deployment choices while holding the underlying model configuration fixed. We consider two such choices: weight precision and hardware platform.

\paragraph{FP8 reduces energy by about a quarter with little change in score}
Across the 28 matched pairs, moving from BF16 to FP8 changes the suite score by a median of \(-0.15\) points (mean \(-0.16\); 95\% CI \(-0.37\) to \(+0.05\)). The corresponding GPU energy falls to \(0.73\times\) that of BF16 (geometric mean; 95\% CI \(0.69\)--\(0.77\)), with 27 of the 28 pairs using less energy.

The saving varies by platform: FP8 uses \(0.69\times\) the BF16 energy on the RTX PRO~6000, \(0.79\times\) on the 8$\times$H100, and \(0.84\times\) on the RTX~5090. The absolute saving consequently grows with model scale, from less than 0.5\,kWh per suite pass for 3--4B models on the RTX~5090 to 7.8\,kWh for Qwen3-235B-A22B-Thinking on the 8$\times$H100. Overall, the energy reduction is much larger and more consistent than the change in capability. This comparison also bounds the weight-format difference in Fig.~\ref{fig:frontier_row}, where the 32 and 96\,GiB panels use BF16 and the 512\,GiB panel uses FP8.

\paragraph{Hardware rescales energy but barely changes the relative efficiency of models}
The same serving configurations have almost identical energy rankings across the four platforms: relative to the RTX PRO~6000, Spearman correlation is \(\rho=0.99\) on every platform, while benchmark scores differ by a median of only 0.4--0.7 points. The score--energy Pareto frontier also contains the same members on every platform, to within one model. Absolute energy nevertheless changes substantially. Depending on weight format, the GH200 uses \(0.61\)--\(0.77\times\) the accelerator energy of the RTX PRO~6000, the GB10 \(0.53\)--\(0.62\times\), and the two RTX A6000s \(1.9\)--\(2.2\times\). The difference is mainly runtime for the GH200, PRO~6000 and A6000 pair; the GB10 instead trades much lower power for substantially longer runtime.

\begin{figure}[ht]
  \centering
  \includegraphics[width=\linewidth]{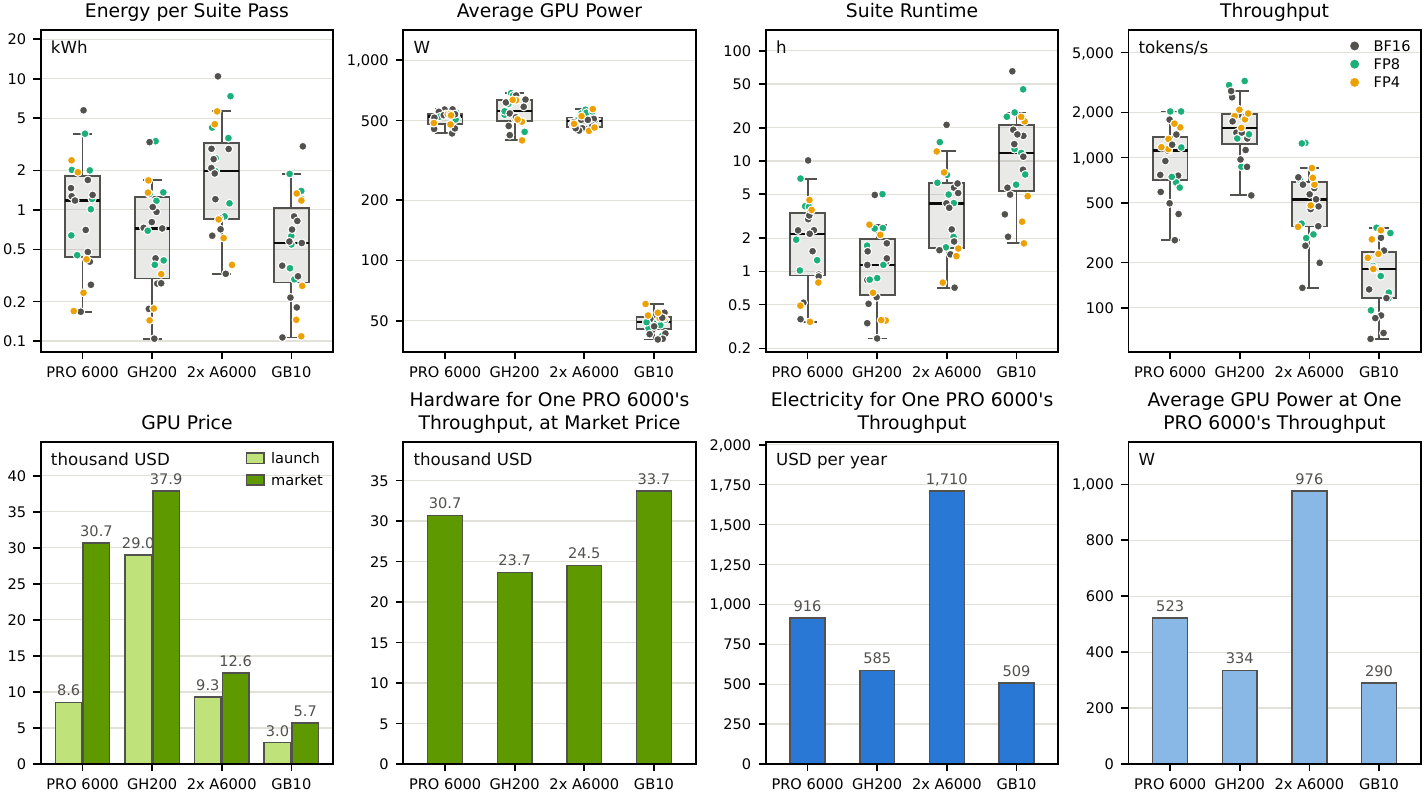}
  \vskip -0.1in

  \caption{\textbf{Energy, performance, and cost across the four 96\,GiB platforms.} \emph{Top:} suite GPU energy, accelerator power, runtime, and token throughput. \emph{Bottom:} accelerator price, hardware cost at matched RTX PRO~6000 throughput, and corresponding annual accelerator electricity cost and power. Secondary-market prices use ten eBay listings; \$1,000 is deducted from the GB10 price for its non-GPU resources. Electricity assumes \$0.20/kWh and continuous operation. Host and infrastructure power are excluded.}
  
  \vskip -0.1in
  \label{fig:xplat_96g}
\end{figure}

\paragraph{Purchase cost dominates accelerator electricity at matched throughput}
At the throughput of one RTX PRO~6000, the required accelerator hardware costs roughly \$24--34k at our indicative secondary-market prices, compared with \$0.5--1.7k per year in accelerator electricity at \$0.20/kWh (Fig.~\ref{fig:xplat_96g}). This is illustrative rather than a full TCO analysis: host systems, cooling, maintenance, and infrastructure are excluded.

\section{Limitations and Scope}

\paragraph{Output budgets}
Scores are conditional on the benchmark-specific generation caps used in our evaluation. Verbose reasoning models can be penalised when answers are truncated: increasing the cap beyond 32K recovers substantial capability for some models, but costs up to \(6.1\times\) more energy and can change their ranking (Fig.~\ref{fig:overthink}). We also report a clean-completion upper bound (Fig.~\ref{fig:frontier_stack_clean}).

\paragraph{Model coverage}
Our largest tier uses a single 8$\times$H100 node and includes models up to 400B parameters. Larger open-weight models were beyond our compute budget, while closed API models fall outside our direct measurement setup.

\paragraph{Measurement and serving}
Our primary measurements cover accelerator energy rather than complete system or data-centre energy; package-level measurements for the GB10 and GH200 are reported in Appendix~\ref{app:energy}. All main experiments use saturated continuous batching, the energy values describe a high-utilisation serving regime and should not be assumed to hold under lighter loads.

\section{Conclusion}
We directly measured the inference energy of 136 serving configurations of open-weight LLMs across seven benchmarks and six hardware platforms under saturated serving. The capability--energy trade-off varies widely: configurations off the Pareto frontier use a median of \(2.6\)--\(4.8\times\) the energy of an alternative that scores at least as well, and up to \(27\times\). Reasoning can improve capability, but at a substantial energy cost, and its benefit depends strongly on the model lineage and output budget; in every memory tier, a non-thinking configuration comes within five points of the best score for only a quarter to a sixth of its energy. Generated-token and parameter counts do not reliably recover measured energy, while FP8 reduces energy by roughly a quarter with little change in score. Changing hardware substantially rescales absolute energy but largely preserves the relative efficiency of models. These results suggest that benchmark capability should be reported together with directly measured energy under a clearly stated serving regime and output budget.

\subsection*{AI use statement}

We used generative AI tools to assist with literature discovery, parts of code development, and
language editing during the preparation of this work. All cited papers were checked against their
original sources, AI-assisted code was reviewed and tested by the authors, and all AI-assisted
writing was manually revised and verified. The authors take full responsibility for the methodology,
experiments, results, and final text.

\subsection*{Ethics statement}

This work evaluates open-weight language models on existing benchmarks and does not involve human
subjects or the collection of personal data. The experiments consume substantial computational
resources and electricity; we therefore report the energy consumed by the evaluation campaign in
Appendix~\ref{app:study_energy}. We are not aware of other significant ethical risks specific to
the methodology or results presented here.

\subsection*{Reproducibility statement}

We document the model roster, sampling configurations, benchmarks, hardware and software
environments, serving regime, energy-measurement procedure, and scoring methodology in the main
paper and Appendix~\ref{app:setup}. These details are intended to support independent reproduction
of the experiments. We plan to release the evaluation codebase and result dataset publicly upon
acceptance.


\bibliography{references}
\bibliographystyle{preprint}

\clearpage

\appendix
\section{Experimental Setup}\label{app:setup}

\subsection{Models}\label{app:models}

Tables~\ref{tab:roster-32g}--\ref{tab:roster-512g} list the checkpoints in each
memory tier. A checkpoint is assigned to the smallest tier in which its stored
weights fit while retaining headroom for the KV cache. Vendor-published FP8 and
NVFP4 releases are treated as separate checkpoints, so lower-precision variants
can appear in a smaller tier than their higher-precision counterparts.

\begin{table}[htbp]
\centering
\footnotesize
\setlength{\tabcolsep}{3pt}
\caption{Roster of the 32\,GiB tier (31 checkpoints, 47 study rows), served on a
single NVIDIA GeForce RTX~5090. \textbf{Params}: billions of parameters; MoE
checkpoints report total\,/\,active parameters (Gemma~4 E-series:
total\,/\,effective). \textbf{Weights}: size of the safetensors weights loaded by
the server. \textbf{Precision}: stored weight format; gpt-oss stores its MoE
experts in MXFP4. \textbf{Modes}: I = instruct, T = thinking, NT = non-thinking;
hybrid checkpoints listed as T,\,NT contribute two study rows. L,\,M,\,H denote
documented low, medium, and high reasoning-effort settings.
\textbf{Released}: checkpoint release date.}
\label{tab:roster-32g}
\begin{tabular}{@{}lrrlll@{}}
\toprule
Checkpoint & Params & Weights & Precision & Modes & Released \\
 & (B) & (GiB) & & & \\
\midrule
\multicolumn{6}{@{}l}{\textbf{Alibaba} \quad \texttt{Qwen/}} \\
\texttt{Qwen3-4B-Instruct-2507} & 4 & 7.5 & BF16 & I & 2025-09-17 \\
\texttt{Qwen3-4B-Instruct-2507-FP8} & 4 & 4.8 & FP8 & I & 2025-09-17 \\
\texttt{Qwen3-4B-Thinking-2507} & 4 & 7.5 & BF16 & T & 2025-08-06 \\
\texttt{Qwen3-4B-Thinking-2507-FP8} & 4 & 4.8 & FP8 & T & 2025-08-06 \\
\texttt{Qwen3.5-0.8B} & 0.8 & 1.6 & BF16 & NT & 2026-03-09 \\
\texttt{Qwen3.5-2B} & 2 & 4.2 & BF16 & T, NT & 2026-03-09 \\
\texttt{Qwen3.5-4B} & 4 & 8.7 & BF16 & T, NT & 2026-03-09 \\
\texttt{Qwen3.5-9B} & 9 & 18.0 & BF16 & T, NT & 2026-03-09 \\
\addlinespace[3pt]
\multicolumn{6}{@{}l}{\textbf{DeepSeek} \quad \texttt{deepseek-ai/}} \\
\texttt{DeepSeek-R1-Distill-Qwen-1.5B} & 1.5 & 3.3 & BF16 & T & 2025-01-20 \\
\texttt{DeepSeek-R1-Distill-Qwen-7B} & 7 & 14.2 & BF16 & T & 2025-01-20 \\
\texttt{DeepSeek-R1-Distill-Llama-8B} & 8 & 15.0 & BF16 & T & 2025-01-20 \\
\addlinespace[3pt]
\multicolumn{6}{@{}l}{\textbf{Google} \quad \texttt{google/}} \\
\texttt{gemma-3-4b-it} & 4 & 8.0 & BF16 & I & 2025-03-12 \\
\texttt{gemma-3-12b-it} & 12 & 22.7 & BF16 & I & 2025-03-12 \\
\texttt{gemma-4-E2B-it} & 5\,/\,2 & 9.5 & BF16 & T, NT & 2026-04-02 \\
\texttt{gemma-4-E4B-it} & 8\,/\,4 & 14.9 & BF16 & T, NT & 2026-04-02 \\
\texttt{gemma-4-12B-it} & 12 & 22.3 & BF16 & T, NT & 2026-04-02 \\
\addlinespace[3pt]
\multicolumn{6}{@{}l}{\textbf{IBM} \quad \texttt{ibm-granite/}} \\
\texttt{granite-4.2-3b} & 3 & 6.8 & BF16 & T, NT & 2026-08-07 \\
\texttt{granite-4.2-8b} & 8 & 16.4 & BF16 & T, NT & 2026-08-07 \\
\texttt{granite-4.2-30b-nvfp4} & 30 & 16.4 & NVFP4 & T, NT & 2026-08-13 \\
\addlinespace[3pt]
\multicolumn{6}{@{}l}{\textbf{Meta} \quad \texttt{meta-llama/}} \\
\texttt{Llama-3.1-8B-Instruct} & 8 & 15.0 & BF16 & I & 2024-07-23 \\
\addlinespace[3pt]
\multicolumn{6}{@{}l}{\textbf{Microsoft} \quad \texttt{microsoft/}} \\
\texttt{Phi-4-mini-instruct} & 3.8 & 7.2 & BF16 & I & 2025-04-30 \\
\addlinespace[3pt]
\multicolumn{6}{@{}l}{\textbf{Mistral AI} \quad \texttt{mistralai/}} \\
\texttt{Ministral-3-3B-Instruct-2512-BF16} & 3 & 7.2 & BF16 & I & 2025-12-02 \\
\texttt{Ministral-3-3B-Instruct-2512} & 3 & 4.3 & FP8 & I & 2025-12-02 \\
\texttt{Ministral-3-3B-Reasoning-2512} & 3 & 7.2 & BF16 & T & 2025-12-02 \\
\texttt{Ministral-3-8B-Instruct-2512} & 8 & 9.7 & FP8 & I & 2025-12-02 \\
\addlinespace[3pt]
\multicolumn{6}{@{}l}{\textbf{NVIDIA} \quad \texttt{nvidia/}} \\
\texttt{Nemotron-3-Nano-4B-BF16} & 4 & 7.4 & BF16 & T, NT & 2026-03-16 \\
\texttt{Nemotron-3-Nano-4B-FP8} & 4 & 4.9 & FP8 & T, NT & 2026-03-16 \\
\texttt{Nemotron-3-Nano-30B-A3B-NVFP4} & 30\,/\,3 & 18.0 & NVFP4 & T, NT & 2025-12-15 \\
\texttt{Nemotron-3-Nano-Omni-30B-A3B-NVFP4} & 30\,/\,3 & 20.9 & NVFP4 & T, NT & 2026-04-28 \\
\texttt{Nemotron-3.5-Lightning-30B-A3B-NVFP4} & 30\,/\,3 & 20.1 & NVFP4 & T, NT & 2026-08-04 \\
\addlinespace[3pt]
\multicolumn{6}{@{}l}{\textbf{OpenAI} \quad \texttt{openai/}} \\
\texttt{gpt-oss-20b} & 21\,/\,3.6 & 12.8 & MXFP4 & L, M, H & 2025-08-05 \\
\bottomrule
\end{tabular}
\end{table}

Hybrid checkpoints contribute one study row per documented reasoning mode,
selected through the vendor-provided chat-template controls (for example,
\texttt{enable\_thinking}). gpt-oss and GLM-5.3-Flash similarly contribute one
row per documented reasoning-effort level. A model-specific reasoning parser in
the serving engine removes the reasoning trace before answer scoring.

Sampling follows vendor-provided settings. For 79 configurations, parameters are
read from the checkpoint's \texttt{generation\_config.json} and applied by the
proxy; for another 17, the serving engine applies that configuration directly.
The 15 Mistral configurations use model-card settings. For the remaining 25
configurations, for which no settings are documented, we use temperature 0.6
and top-$p$ 0.95.

\begin{table}[htbp]
\centering
\footnotesize
\setlength{\tabcolsep}{3pt}
\caption{Roster of the 96\,GiB tier (29 checkpoints, 47 study rows), served on
all four 96\,GiB platforms: GB10, RTX PRO~6000 Blackwell, GH200, and
2$\times$RTX~A6000 (TP=2). Columns are defined in
Table~\ref{tab:roster-32g}.}
\label{tab:roster-96g}
\begin{tabular}{@{}lrrlll@{}}
\toprule
Checkpoint & Params & Weights & Precision & Modes & Released \\
 & (B) & (GiB) & & & \\
\midrule
\multicolumn{6}{@{}l}{\textbf{Alibaba} \quad \texttt{Qwen/}} \\
\texttt{Qwen3-30B-A3B-Instruct-2507} & 30\,/\,3 & 56.9 & BF16 & I & 2025-09-17 \\
\texttt{Qwen3-30B-A3B-Instruct-2507-FP8} & 30\,/\,3 & 29.0 & FP8 & I & 2025-09-17 \\
\texttt{Qwen3-30B-A3B-Thinking-2507} & 30\,/\,3 & 56.9 & BF16 & T & 2025-08-17 \\
\texttt{Qwen3-30B-A3B-Thinking-2507-FP8} & 30\,/\,3 & 29.0 & FP8 & T & 2025-07-30 \\
\texttt{Qwen3.6-27B} & 27 & 51.8 & BF16 & T, NT & 2026-04-24 \\
\texttt{Qwen3.6-27B-FP8} & 27 & 28.8 & FP8 & T, NT & 2026-04-24 \\
\texttt{Qwen3.6-35B-A3B} & 35\,/\,3 & 67.0 & BF16 & T, NT & 2026-04-24 \\
\texttt{Qwen3.6-35B-A3B-FP8} & 35\,/\,3 & 34.9 & FP8 & T, NT & 2026-04-24 \\
\texttt{Qwen3.8-27B} & 27 & 51.8 & BF16 & T, NT & 2026-08-05 \\
\texttt{Qwen3.8-27B-FP8} & 27 & 28.8 & FP8 & T, NT & 2026-08-13 \\
\addlinespace[3pt]
\multicolumn{6}{@{}l}{\textbf{DeepSeek} \quad \texttt{deepseek-ai/}} \\
\texttt{DeepSeek-R1-Distill-Qwen-14B} & 14 & 27.5 & BF16 & T & 2025-01-20 \\
\texttt{DeepSeek-R1-Distill-Qwen-32B} & 32 & 61.0 & BF16 & T & 2025-01-20 \\
\addlinespace[3pt]
\multicolumn{6}{@{}l}{\textbf{Google} \quad \texttt{google/}} \\
\texttt{gemma-3-27b-it} & 27 & 51.1 & BF16 & I & 2025-03-12 \\
\texttt{gemma-4-26B-A4B-it} & 26\,/\,4 & 48.1 & BF16 & T, NT & 2026-04-02 \\
\texttt{gemma-4-31B-it} & 31 & 58.2 & BF16 & T, NT & 2026-04-02 \\
\addlinespace[3pt]
\multicolumn{6}{@{}l}{\textbf{IBM} \quad \texttt{ibm-granite/}} \\
\texttt{granite-4.2-30b} & 30 & 54.5 & BF16 & T, NT & 2026-08-07 \\
\texttt{granite-4.2-30b-fp8} & 30 & 28.0 & FP8 & T, NT & 2026-08-13 \\
\addlinespace[3pt]
\multicolumn{6}{@{}l}{\textbf{Mistral AI} \quad \texttt{mistralai/}} \\
\texttt{Ministral-3-8B-Instruct-2512-BF16} & 8 & 16.6 & BF16 & I & 2025-12-02 \\
\texttt{Ministral-3-8B-Reasoning-2512} & 8 & 16.6 & BF16 & T & 2025-12-02 \\
\texttt{Ministral-3-14B-Instruct-2512-BF16} & 14 & 26.0 & BF16 & I & 2025-12-02 \\
\texttt{Ministral-3-14B-Instruct-2512} & 14 & 14.7 & FP8 & I & 2025-12-02 \\
\texttt{Ministral-3-14B-Reasoning-2512} & 14 & 26.0 & BF16 & T & 2025-12-02 \\
\texttt{Mistral-Small-4-119B-2603-NVFP4} & 119\,/\,6.5 & 66.0 & NVFP4 & T, NT & 2026-03-16 \\
\addlinespace[3pt]
\multicolumn{6}{@{}l}{\textbf{NVIDIA} \quad \texttt{nvidia/}} \\
\texttt{Nemotron-3-Nano-30B-A3B-BF16} & 30\,/\,3 & 58.8 & BF16 & T, NT & 2025-12-15 \\
\texttt{Nemotron-3-Nano-30B-A3B-FP8} & 30\,/\,3 & 30.4 & FP8 & T, NT & 2025-12-15 \\
\texttt{Nemotron-3-Super-120B-A12B-NVFP4} & 120\,/\,12 & 74.8 & NVFP4 & T, NT & 2026-03-11 \\
\texttt{Nemotron-3.5-Lightning-30B-A3B-BF16} & 30\,/\,3 & 61.3 & BF16 & T, NT & 2026-08-01 \\
\addlinespace[3pt]
\multicolumn{6}{@{}l}{\textbf{OpenAI} \quad \texttt{openai/}} \\
\texttt{gpt-oss-120b} & 117\,/\,5.1 & 60.8 & MXFP4 & L, M, H & 2025-08-05 \\
\addlinespace[3pt]
\multicolumn{6}{@{}l}{\textbf{Z.ai} \quad \texttt{zai-org/}} \\
\texttt{GLM-4.7-Flash} & 30\,/\,3 & 58.2 & BF16 & T, NT & 2026-01-14 \\
\bottomrule
\end{tabular}
\end{table}

\begin{table}[htbp]
\centering
\footnotesize
\setlength{\tabcolsep}{3pt}
\caption{Roster of the 512\,GiB tier (26 checkpoints, 42 study rows), served on
the 8$\times$H100 node with TP=8 and expert parallelism enabled for MoE
checkpoints. Columns are defined in Table~\ref{tab:roster-32g}.}
\label{tab:roster-512g}
\begin{tabular}{@{}lrrlll@{}}
\toprule
Checkpoint & Params & Weights & Precision & Modes & Released \\
 & (B) & (GiB) & & & \\
\midrule
\multicolumn{6}{@{}l}{\textbf{Alibaba} \quad \texttt{Qwen/}} \\
\texttt{Qwen3-235B-A22B-Instruct-2507} & 235\,/\,22 & 437.9 & BF16 & I & 2025-09-17 \\
\texttt{Qwen3-235B-A22B-Instruct-2507-FP8} & 235\,/\,22 & 220.2 & FP8 & I & 2025-09-17 \\
\texttt{Qwen3-235B-A22B-Thinking-2507} & 235\,/\,22 & 437.9 & BF16 & T & 2025-08-17 \\
\texttt{Qwen3-235B-A22B-Thinking-2507-FP8} & 235\,/\,22 & 220.2 & FP8 & T & 2025-07-30 \\
\texttt{Qwen3.5-122B-A10B} & 122\,/\,10 & 233.0 & BF16 & T, NT & 2026-03-09 \\
\texttt{Qwen3.5-122B-A10B-FP8} & 122\,/\,10 & 118.4 & FP8 & T, NT & 2026-03-09 \\
\texttt{Qwen3.5-397B-A17B-FP8} & 397\,/\,17 & 378.3 & FP8 & T, NT & 2026-03-09 \\
\texttt{Qwen3.8-Flash-Next}     & 180\,/\,6 & 335.3 & BF16 & T, NT & 2026-08-24 \\
\texttt{Qwen3.8-Flash-Next-FP8} & 180\,/\,6 & 172.8 & FP8 & T, NT & 2026-08-24 \\

\addlinespace[3pt]
\multicolumn{6}{@{}l}{\textbf{DeepSeek} \quad \texttt{deepseek-ai/}} \\
\texttt{DeepSeek-R1-Distill-Llama-70B} & 70 & 131.4 & BF16 & T & 2025-01-20 \\
\texttt{DeepSeek-V4-Flash} & 284\,/\,13 & 148.7 & FP4/FP8 & T, NT & 2026-04-24 \\

\addlinespace[3pt]
\multicolumn{6}{@{}l}{\textbf{Meta} \quad \texttt{meta-llama/}} \\
\texttt{Llama-3.1-70B-Instruct} & 70 & 131.4 & BF16 & I & 2024-07-23 \\
\texttt{Llama-4-Scout-17B-16E-Instruct} & 109\,/\,17 & 202.4 & BF16 & I & 2025-04-05 \\
\texttt{Llama-4-Maverick-17B-128E-Instruct} & 400\,/\,17 & 388.1 & FP8 & I & 2025-04-05 \\
\addlinespace[3pt]
\multicolumn{6}{@{}l}{\textbf{MiniMax} \quad \texttt{MiniMaxAI/}} \\
\texttt{MiniMax-M2} & 230\,/\,10 & 214.3 & FP8 & T & 2025-10-27 \\
\texttt{MiniMax-M2.5} & 230\,/\,10 & 214.3 & FP8 & T & 2026-02-12 \\
\texttt{MiniMax-M2.7} & 230\,/\,10 & 214.3 & FP8 & T & 2026-03-18 \\
\addlinespace[3pt]
\multicolumn{6}{@{}l}{\textbf{Mistral AI} \quad \texttt{mistralai/}} \\
\texttt{Mistral-Medium-3.5-128B} & 128 & 248.9 & BF16 & T, NT & 2026-04-29 \\
\texttt{Mistral-Small-4-119B-2603} & 119\,/\,6.5 & 225.2 & BF16 & T, NT & 2026-03-16 \\
\addlinespace[3pt]
\multicolumn{6}{@{}l}{\textbf{NVIDIA} \quad \texttt{nvidia/}} \\
\texttt{Nemotron-3-Super-120B-A12B-BF16} & 120\,/\,12 & 230.2 & BF16 & T, NT & 2026-03-11 \\
\texttt{Nemotron-3-Super-120B-A12B-FP8} & 120\,/\,12 & 119.5 & FP8 & T, NT & 2026-03-11 \\
\addlinespace[3pt]
\multicolumn{6}{@{}l}{\textbf{Z.ai} \quad \texttt{zai-org/}} \\
\texttt{GLM-4.5-Air} & 106\,/\,12 & 205.8 & BF16 & T, NT & 2025-07-28 \\
\texttt{GLM-4.5-Air-FP8} & 106\,/\,12 & 104.8 & FP8 & T, NT & 2025-07-28 \\
\texttt{GLM-4.5-FP8} & 355\,/\,32 & 336.5 & FP8 & T, NT & 2025-07-28 \\
\texttt{GLM-4.7-FP8} & 355\,/\,32 & 337.2 & FP8 & T, NT & 2025-12-22 \\
\texttt{GLM-5.3-Flash} & 320\,/\,18 & 305.8 & FP8 & L, H, M & 2026-08-25 \\

\bottomrule
\end{tabular}
\end{table}

\newpage

\subsection{Hardware}\label{app:hardware}

Table~\ref{tab:platform-details} gives the hardware, software, measurement, and
serving details of each platform. All accelerators run at their default power
limits. Single-accelerator platforms serve each model on one device; the A6000
pair uses tensor parallelism over NVLink, while the 8$\times$H100 node uses
tensor parallelism across eight GPUs with expert parallelism enabled for MoE
models. The GB10 serving engine is limited to 96\,GiB of its 128\,GB unified
memory.

The server admission limit follows the pinned serving engine's default: 256
concurrent sequences on the RTX~5090, A6000 pair, and GB10, and up to 1{,}024
on the other three platforms, with the engine reducing the limit where required
by the model.

\begin{table}[htbp]
\centering
\footnotesize
\setlength{\tabcolsep}{3pt}
\renewcommand{\arraystretch}{1.1}
\caption{Hardware, software, measurement, and serving details for the six platforms in Table~\ref{tab:platforms}.}
\label{tab:platform-details}
\begin{tabularx}{\textwidth}{@{}l*{6}{>{\raggedright\arraybackslash}X}@{}}
\toprule
 & GeForce RTX 5090 & 2$\times$ RTX A6000 & GH200 & GB10 (DGX Spark) & RTX PRO 6000 Blackwell & 8$\times$ H100 SXM \\
\midrule
\multicolumn{7}{@{}l}{\textit{\textbf{\underline{System}}}} \\
Host CPU & Ryzen 9 7900X & Xeon w7-3455 & Grace & 10$\times$ Cortex-X925 + 10$\times$ Cortex-A725 & Core i5-14600K & 2$\times$ Xeon Platinum 8468 \\
Cores/Threads & 12C/24T & 24C/48T & 72C/72T & 20C/20T & 6P8E/20T & 96C/192T \\
Host RAM & 64 GB DDR5 & 256 GB DDR5 & 480\,GB LPDDR5X & 128 GB unified LPDDR5X & 128 GB DDR5 & 1 TB DDR5 \\
OS & Ubuntu 24.04 & Ubuntu 24.04 & AlmaLinux 9.4 & Ubuntu 24.04 & Ubuntu 22.04 & AlmaLinux 9.4 \\
kernel & 6.8 & 6.8 & 5.14 & 6.14 & 6.8 & 5.14 \\
Driver & 580.126 & 580.65 & 580.65 & 580.95 & 595.84 & 580.65 \\
CUDA &  13.0 & 13.0 & 13.0 & 13.0 & 13.2 & 13.0 \\
Power limit (W) & 575 (default) & 300 (each) & 900 & 140 (SoC) & 600 & 700 (each) \\
\addlinespace
Meter Scope & board & 2 boards, summed & GPU+HBM (module recorded) & GPU rail (package recorded) & board & 8 boards, summed \\
\addlinespace
\multicolumn{7}{@{}l}{\textit{\textbf{\underline{Serving}}}} \\
Parallelism & single GPU & TP=2 (NVLink) & single GPU & single GPU & single GPU & TP=8 (NVSwitch); MoE: EP=8 \\
Memory budget & 0.95 & 0.95 & 0.95 & 96.0\,GiB & 0.95 & 0.95 \\
Attention & FA2 & FA2 & FA3 & FA2 & FA2 & FA3 \\
FP8 kernel & native & Marlin W8A16 & native & native & native & native \\
NVFP4 kernel & native & Marlin W4A16 & Marlin W4A16 & native & native & Marlin W4A16 \\
\bottomrule
\end{tabularx}
\end{table}

\clearpage
\subsection{Energy Measurement}\label{app:energy}

\paragraph{Instrumentation}
On discrete GPUs, board power is read from the driver's onboard sensor and
integrated over time; multi-GPU energy is summed across devices. On the GB10,
we read the SoC's cumulative energy counters and use the GPU-rail counter as the
primary measurement, while also recording the CPU and package counters. On the
GH200, the primary GPU reading covers the GPU and HBM3; whole-module power for
the Grace Hopper superchip is recorded alongside it.

\begin{figure}[ht]
  \centering
  \vskip -0.05in
  \includegraphics[width=\linewidth]{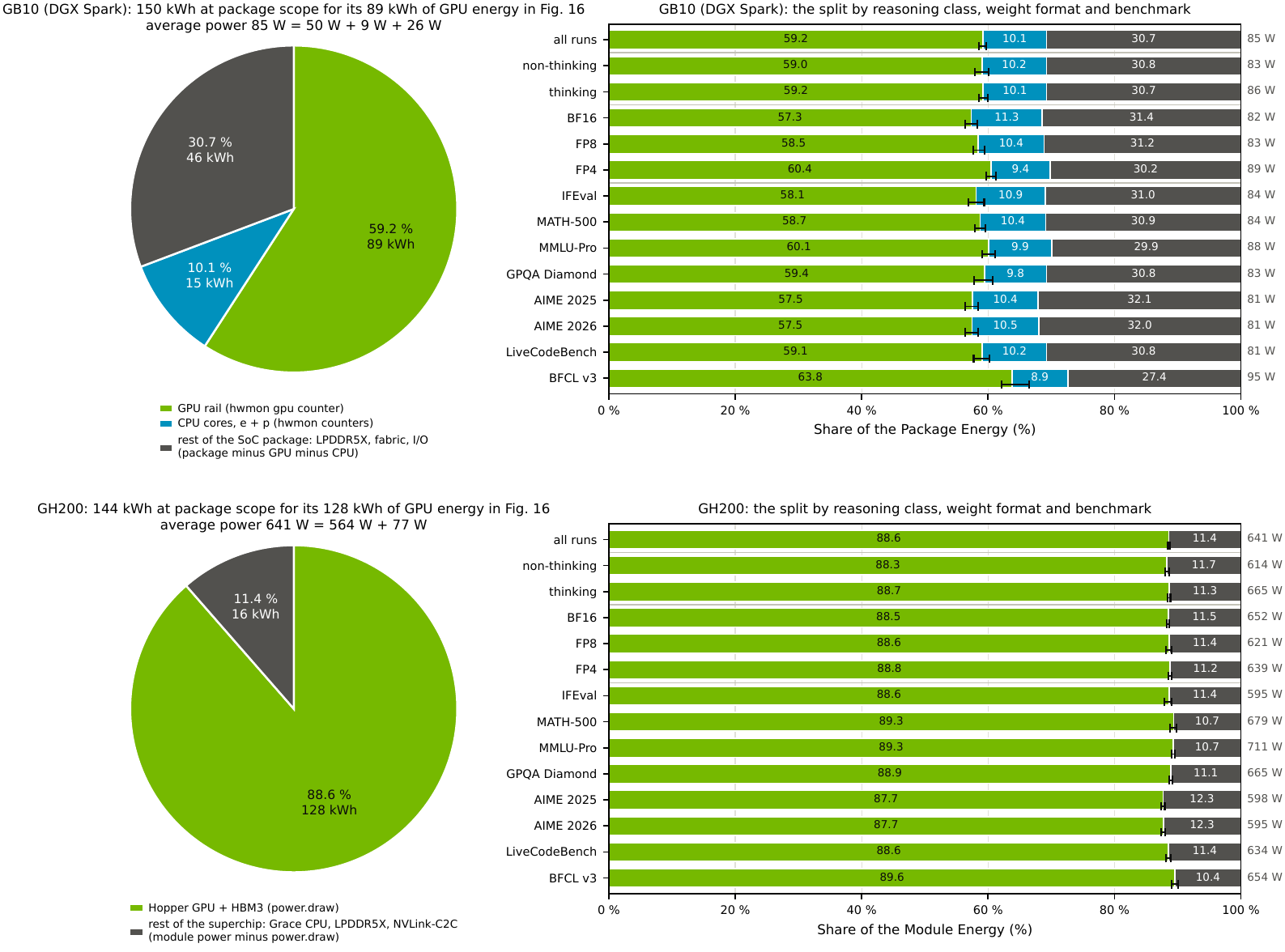}
  \vskip -0.05in
\caption{\textbf{Energy outside the primary accelerator measurement on the two
superchips.} \emph{Left:} energy over all metered attempts, split by component.
For the GB10, the primary measurement is the
GPU rail; for the GH200, it is the
GPU+HBM reading. \emph{Right:} the same shares by reasoning mode, weight format,
and benchmark, with 95\% bootstrap intervals on the primary-counter share and
mean power at right. Package/module measurements still exclude power-supply
losses, fans, and storage.}
  \vskip -0.05in
  \label{fig:soc_power_share}
\end{figure}

\paragraph{Integrated superchips}
Figure~\ref{fig:soc_power_share} compares these primary accelerator measurements
with the wider package or module boundary. On the GB10, package energy is a
median \(1.69\times\) the GPU-rail energy (5th--95th percentile
\(1.52\)--\(1.85\times\)): 59\% is attributed to the GPU rail, 10\% to the CPU
cores, and 31\% to the remainder of the package, including the LPDDR5X memory.
On the GH200, module energy is \(1.125\times\) the GPU+HBM reading
(\(1.11\)--\(1.19\times\)); the remaining 11\% covers the Grace CPU, LPDDR5X,
and NVLink-C2C.

The additional package or module draw is approximately constant at 34\,W on the
GB10 and 77\,W on the GH200, so these shares vary little across reasoning modes,
weight formats, and benchmarks. The measurement boundary matters when comparing
against discrete GPUs, whose board-power measurements include device memory.
For the same configurations, the GB10 consumes \(0.61\times\) the RTX
PRO~6000 energy at GPU scope but \(1.03\times\) at package scope; the
corresponding GH200 ratios are \(0.64\times\) and \(0.73\times\).

\clearpage
\subsection{Suite Aggregation}\label{app:aggregation}

The suite contains seven components, with AIME 2025 and AIME 2026 combined into
one component by averaging their scores and summing their energies. Suite score
is the arithmetic mean of the seven component scores, while suite energy is the
geometric mean of their per-attempt energies.

\begin{figure}[ht]
  \centering
  \vskip -0.05in
  \includegraphics[width=\linewidth]{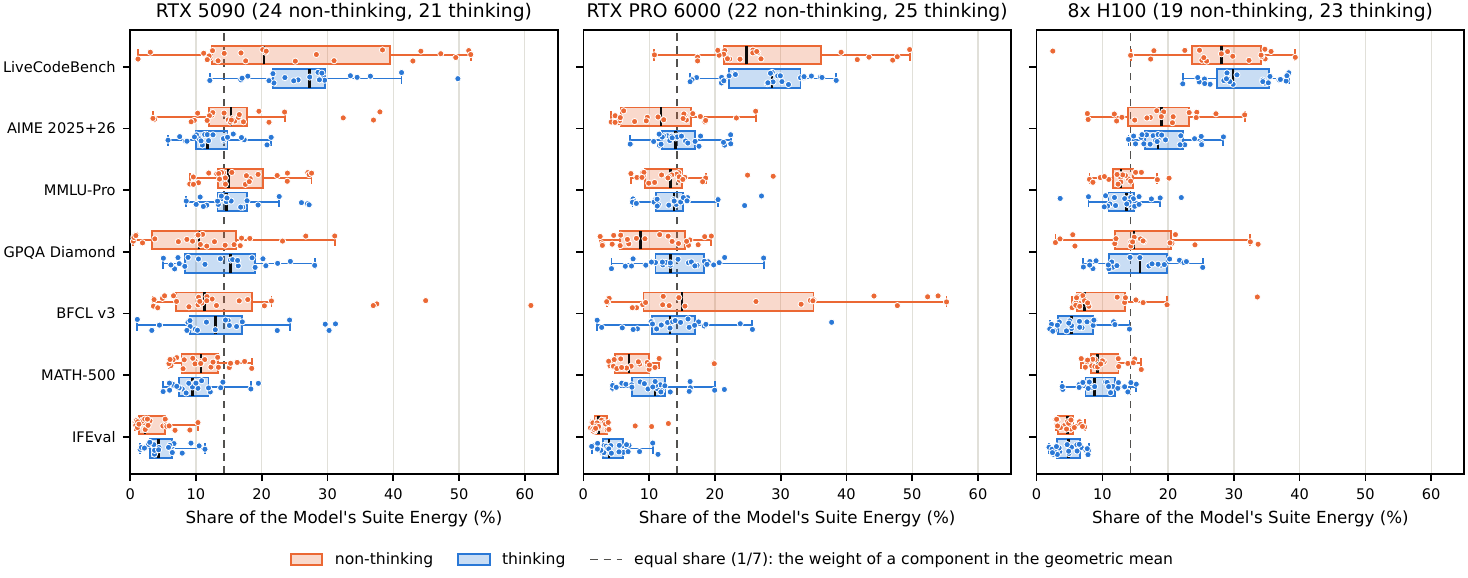}
  \vskip -0.05in
\caption{\textbf{Benchmark shares of summed suite-pass energy.}
For every complete configuration on each reference platform, boxes show the share
of total suite-pass energy contributed by each benchmark, separated into
non-thinking (orange) and thinking (blue) configurations. Points show individual
configurations. The dashed line marks one seventh, the equal weight assigned to
each component by the geometric mean.}
  \vskip -0.05in
  \label{fig:energy_shares}
\end{figure}

\begin{figure}[ht]
  \centering
  \vskip -0.05in
  \includegraphics[width=\linewidth]{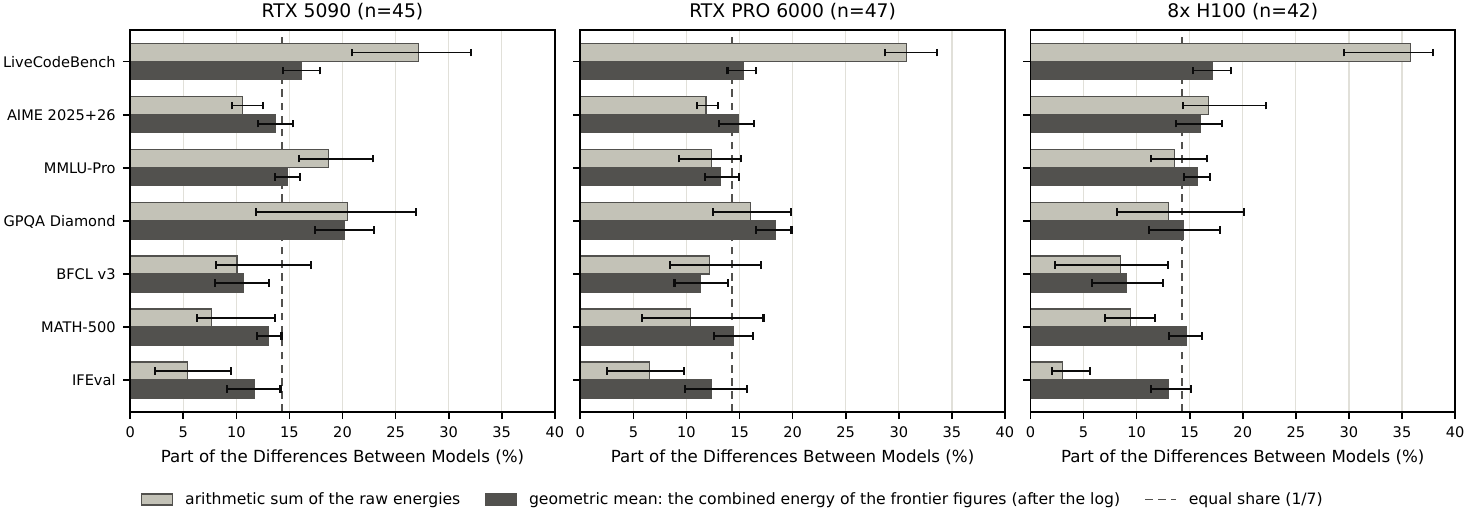}
  \vskip -0.05in
\caption{\textbf{Each benchmark's contribution to cross-configuration variation
in aggregate energy.}
For each reference platform, bars show the contribution of each benchmark under
the arithmetic sum of raw energies (light) and the geometric mean used in the
paper (dark, on log energies). Contributions sum to 100\%; error bars are 95\%
bootstrap intervals over configurations. The dashed line marks an equal
one-seventh contribution.}
  \vskip -0.05in
  \label{fig:energy_contrib}
\end{figure}

We use the geometric mean to prevent benchmarks with more or longer generations
from dominating the aggregate. Under a simple sum, LiveCodeBench contributes a
median 25--29\% of the summed suite-pass energy and is the largest component for
93 of the 134 configurations with complete reference-platform results, whereas
IFEval contributes only 3--5\% (Fig.~\ref{fig:energy_shares}). LiveCodeBench
also accounts for 27--36\% of the cross-configuration variation under the sum,
compared with 3--7\% for IFEval. Under the geometric mean, each component
contributes 9--20\%, much closer to the equal share of one seventh
(Fig.~\ref{fig:energy_contrib}).

The aggregation choice has little effect on model ordering: the arithmetic-sum
and geometric-mean rankings have Spearman \(\rho=0.985\)--\(0.995\), and
dropping any one component gives \(\rho\ge0.986\). We use the geometric mean
because changing a benchmark's number of items or attempts then rescales all
configurations equally, leaving pairwise energy ratios and rankings unchanged.

\clearpage
\section{Additional Results}\label{app:results}

\subsection{The Frontier in Detail}\label{app:frontier}

\begin{figure}[ht]
  \centering
  \vskip -0.05in
  \includegraphics[width=\linewidth]{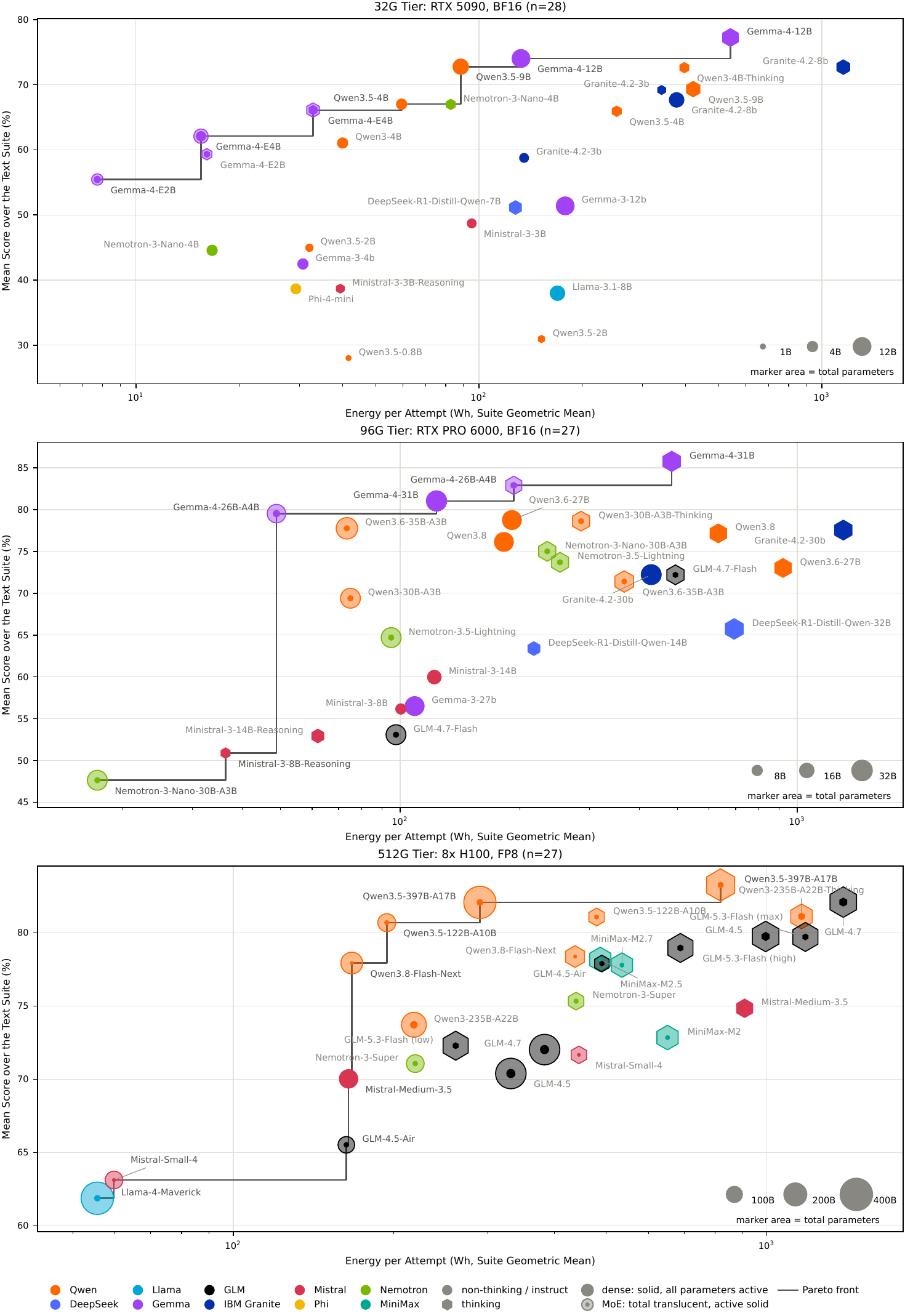}
  \vskip -0.1in
\caption{\textbf{Score--energy fronts with every configuration labelled.}
Enlarged versions of the three panels in Fig.~\ref{fig:frontier_row};
marker encoding is unchanged.}
  \vskip -0.05in
  \label{fig:frontier_stack}
\end{figure}

Fig.~\ref{fig:frontier_stack} reproduces the three panels of
Fig.~\ref{fig:frontier_row} at full width with every configuration labelled.
Each panel uses the majority weight format of its tier: BF16 for the 32 and
96\,GiB tiers, and FP8 for the 512\,GiB tier.

\paragraph{Robustness of the fronts}
We recompute the fronts over 2{,}000 resampled draws. In each draw, benchmark
scores are resampled over their items and attempts, giving a typical suite-score
standard error of about 0.8 points, while component energies are perturbed by
7\%, the median variation between repeated attempts of the same benchmark.
Eighteen of the 21 frontier members remain on their front in at least 85\% of
draws.

The three less-stable members are all near-ties. Qwen3.5-4B without thinking is
within 0.1 score points of Nemotron-3-Nano-4B with thinking and remains on the
front in 77\% of draws. On the 512\,GiB tier, Mistral-Medium-3.5 and
GLM-4.5-Air without thinking are within 2\% of the energy of neighbouring
frontier configurations and remain on the front in 65\% and 49\% of draws,
respectively. No off-front configuration joins a frontier in more than 40\% of
draws. When any one suite component is removed, 16 of the 21 frontier members
remain.

\begin{figure}[ht]
  \centering
  \vskip -0.1in
  \includegraphics[width=\linewidth]{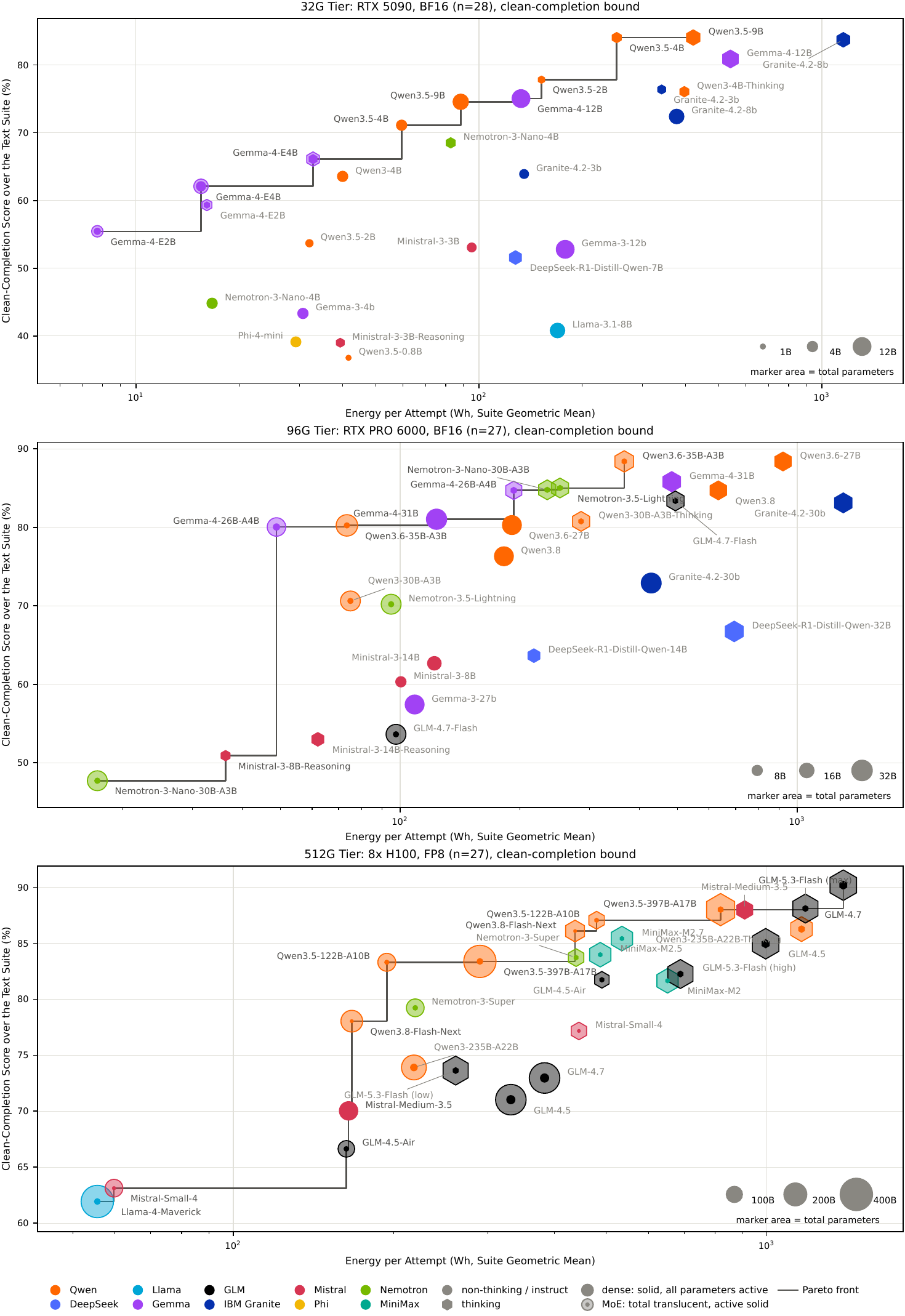}
  \vskip -0.14in
\caption{\textbf{Score--energy fronts under the clean-completion bound.}
As in Fig.~\ref{fig:frontier_stack}, but benchmark scores are recomputed using
only requests that finish within the output cap. Energy is unchanged and the
Pareto fronts are recomputed.}
  \vskip -0.1in
  \label{fig:frontier_stack_clean}
\end{figure}

\paragraph{The clean-completion bound}
Fig.~\ref{fig:frontier_stack_clean} replaces each benchmark score with its score
over requests that complete within the output cap, equivalently dividing the
observed score by one minus the truncated-request fraction when truncated
requests contribute zero.

Energy is left unchanged. The bound is deliberately optimistic: it assumes that
truncated answers would be correct at the same rate as completed answers, while
charging no additional energy for allowing them to finish. It therefore
estimates an upper bound on the capability recoverable from a larger output
budget.

The effect is largest where truncation is most severe. Across the three tiers,
thinking configurations gain a median of 3.5, 3.9, and 5.9 points, compared with
2.2, 0.9, and 0.6 points for non-thinking configurations. Overall rankings remain
similar to the scored suite (Spearman \(\rho=0.79\), 0.75, and 0.85), but
thinking configurations move to the top of each frontier. On the RTX~5090,
Qwen3.5 thinking configurations join the frontier at scores of 78--84; on the
RTX PRO~6000, Qwen3.6-35B-A3B and two Nemotron thinking configurations join at
85--88; and on the 8$\times$H100, four thinking configurations join at 86--90.
Every non-thinking frontier member remains.

\clearpage
\subsection{Energy per Correct Answer}\label{app:per_correct}

\begin{figure}[ht]
  \centering
  \vskip -0.05in
  \includegraphics[width=\linewidth]{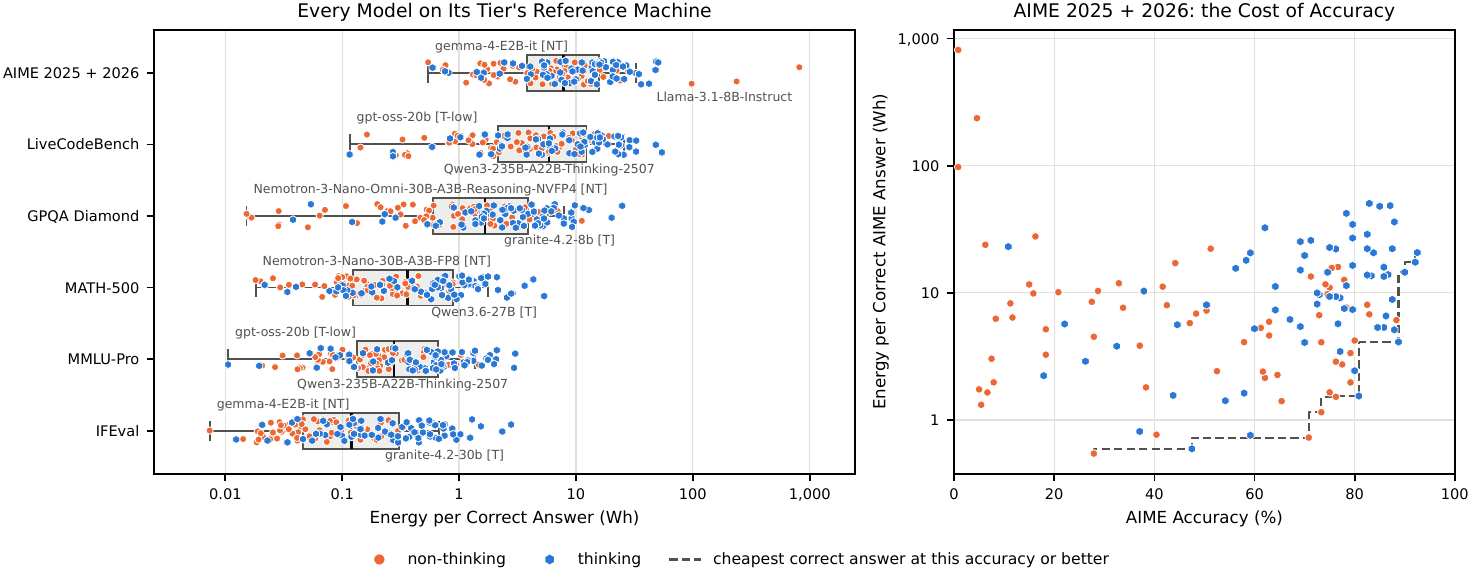}
  \vskip -0.05in
\caption{\textbf{Energy per correct answer.}
All configurations are shown on their tier's reference platform; AIME~2025 and
2026 are pooled, and BFCL is excluded. \emph{Left:} energy per correct answer
across benchmarks; boxes summarise configurations and points denote reasoning
mode, with the extremes labelled. \emph{Right:} energy per correct AIME answer
against AIME accuracy. The dashed step gives the minimum energy per correct
answer among configurations achieving at least each accuracy.}
  \vskip -0.05in
  \label{fig:energy_per_correct}
\end{figure}

Fig.~\ref{fig:energy_per_correct} reports benchmark energy divided by the number
of correct answers for every configuration on its tier's reference platform.
BFCL is excluded because its headline score is a category macro-average rather
than a count of correct answers.

The cost of a correct AIME answer ranges from 0.54 to 815\,Wh. The upper end is
driven by configurations with near-zero accuracy; excluding configurations below
5\% accuracy narrows the range to 0.54--50\,Wh. Across the other benchmarks,
the median cost per correct answer ranges from 0.12\,Wh on IFEval to 5.9\,Wh on
LiveCodeBench, with thinking configurations having a 1.7--4.5$\times$ higher
median than non-thinking configurations.

Cost and accuracy are only weakly correlated, with Spearman
\(\rho=0.12\)--\(0.37\) across benchmarks. Among the 29 configurations achieving
at least 80\% on AIME, a correct answer costs from 1.5\,Wh
(gpt-oss-120b, medium effort, RTX PRO~6000) to 50\,Wh
(GLM-4.7 with thinking, 8$\times$H100), a 33$\times$ spread. Within every
five-point accuracy band from 40\% to 90\%, the most energy-intensive
configuration uses 12--36$\times$ as much energy per correct answer as the
least. These comparisons span memory tiers and therefore combine model and
hardware effects.

\clearpage
\subsection{Efficiency over Release Date}\label{app:release}

\begin{figure}[ht]
  \centering
  \vskip -0.05in
  \includegraphics[width=\linewidth]{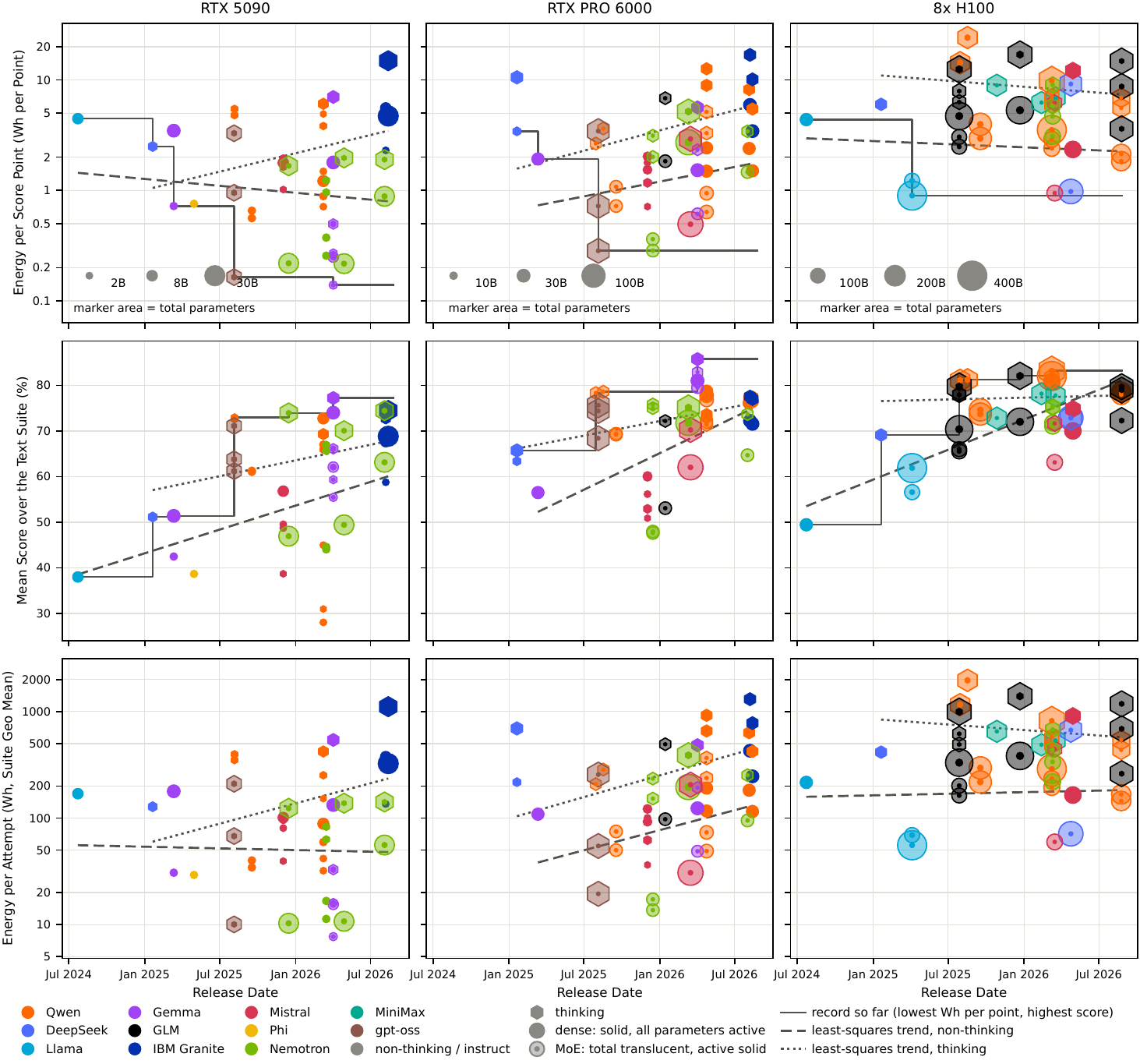}
  \vskip -0.05in
\caption{\textbf{Efficiency, capability, and energy over release date.}
Each complete configuration is placed at its checkpoint release date on its
tier's reference platform, across all weight formats. Rows show energy per score
point, suite score, and suite energy. Solid steps show the best value observed
up to each date; dashed and dotted lines show log-scale least-squares trends for
non-thinking and thinking configurations. Marker encoding follows
Fig.~\ref{fig:frontier_row}.}
  \vskip -0.05in
  \label{fig:efficiency_years}
\end{figure}

Fig.~\ref{fig:efficiency_years} places each complete configuration at its
checkpoint release date. Energy per score point spans \(114\times\),
\(59\times\), and \(27\times\) across the three reference platforms, but shows
no detectable trend with release date: the fitted multiplicative change per year
is \(1.18\times\), \(1.77\times\), and \(1.07\times\), with all 95\%
intervals including one.

Scores do increase with release date. Among non-thinking configurations, the
fitted increase is 22--29\% per year on all three platforms, while the energy of
thinking configurations increases at least as quickly on the two single-GPU
platforms. The best observed energy per score point improves by
\(32\times\), \(12\times\), and \(5\times\) relative to the oldest
configurations, although the current records were already established between
April 2025 and April 2026.

Within lineages at matched model size, the 30 successor transitions change energy
per score point by a median of \(0.97\times\). Non-thinking successors gain
7.6 score points at approximately unchanged energy, whereas thinking successors
lose 1.3 points while using \(0.94\times\) the energy. These temporal trends
should be interpreted cautiously: the roster contains few releases before
mid-2025 and is not a random sample of the model ecosystem.

\clearpage
\section{Serving Outside the Saturated Regime}\label{app:serving}

\subsection{Batch Size 1}\label{app:batch1}
On the 8$\times$H100 node, we re-served the AIME~2026 and LiveCodeBench traffic
of seven configurations with one request in flight, preserving the prompt and
generation lengths of the benchmark requests. After warm-up, each configuration
was metered for 256\,s. The seven configurations comprise three MoE checkpoints
in both reasoning modes and the dense DeepSeek-R1-Distill-Llama-70B.

At batch size~1, the node draws 1.7--1.9\,kW for the MoE configurations and
3.2\,kW for the dense model, while generating 71--190 tokens/s. For a given
configuration, energy per generated token varies by at most \(1.08\times\)
between the two benchmarks and increases with active parameter count, from
9.8\,J at 12B active parameters to 33\,J for the dense 70B model.

Relative to the corresponding campaign runs, energy per generated token is
\(6.6\times\) higher at batch size~1 on average (geometric mean;
\(2.9\)--\(18.5\times\) across configuration--benchmark pairs). Throughput
falls by \(11\times\), while power falls by only \(1.5\times\). These results
cover seven configurations on one platform and quantify the sensitivity to
batching rather than a universal batch-size penalty.

\begin{figure}[ht]
  \centering
  \vskip -0.05in
  \includegraphics[width=\linewidth]{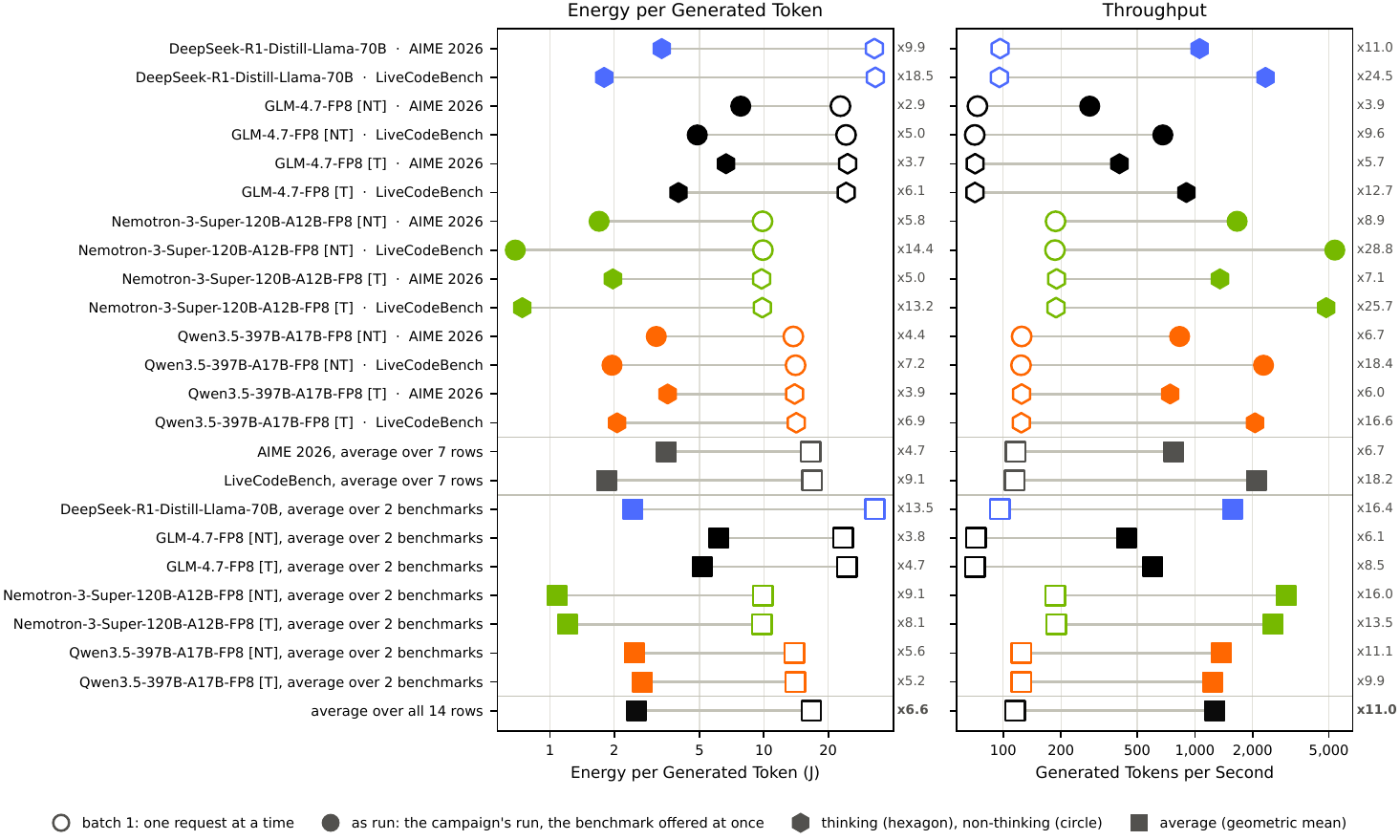}
  \vskip -0.05in
\caption{\textbf{Batch size~1 against the campaign serving regime on the
8$\times$H100 node.}
Energy per generated token (\emph{left}) and throughput (\emph{right}) for seven
configurations on AIME~2026 and LiveCodeBench. Open markers show one request at
a time; filled markers show the corresponding campaign runs, with multiplicative
gaps annotated. Squares show geometric means by benchmark, configuration, and
across all 14 pairs. Colour denotes lineage; circles are non-thinking and
hexagons thinking configurations.}
  \vskip -0.05in
  \label{fig:batch1}
\end{figure}

\subsection{Weights Offloaded to CPU Memory}\label{app:offload}

The models in the 512\,GiB tier do not fit within the GH200's 96\,GiB of HBM3
under our main serving configuration. We therefore serve two of them on a single
GH200 using the serving engine's CPU-offload mechanism, which places weights
that do not fit in HBM3 in Grace LPDDR5X and transfers them over NVLink-C2C
during inference. The two configurations are Qwen3.5-122B-A10B
(68 of 118\,GiB offloaded; 57\%) and Qwen3.5-397B-A17B
(344 of 378\,GiB; 91\%), both FP8 without thinking. We evaluate MMLU-Pro,
AIME~2026, and LiveCodeBench under otherwise unchanged campaign settings.

For comparison, we fit a least-squares trend on log scales against active
parameter count using the seven FP8 non-thinking 96\,GiB configurations whose
weights fit entirely in HBM3 on the same GH200 (3--30B active parameters).
Relative to this trend, the offloaded runs take \(7\)--\(19\times\) longer,
achieve only \(0.04\)--\(0.22\times\) the throughput, and use
\(1.8\)--\(7.2\times\) more energy per generated token. The energy penalty is
smaller than the runtime penalty because GPU+HBM power falls to 200--250\,W
during the offloaded runs, compared with 290--700\,W for the in-memory
configurations. The result is robust to the choice of reference configurations:
throughput remains \(0.05\)--\(0.26\times\) and energy per token
\(1.5\)--\(7.2\times\) the fitted trend.

With 14--131 requests in flight, a uniform-routing approximation implies that
each decode step accesses 27--97\% of the experts, corresponding to roughly
180--330\,GiB/s of offloaded-weight traffic over NVLink-C2C. Together with the
large throughput loss and reduced accelerator power, this is consistent with
offloaded-weight traffic becoming the serving bottleneck.

Offloading changes serving cost much more than capability. Scores differ from
the same models on the 8$\times$H100 node by at most 1.2 points on MMLU-Pro and
LiveCodeBench, and by at most 6 points on the 30-problem AIME~2026. At full module scope, including the Grace CPU and
memory holding the offloaded weights, energy rises to \(1.4\)--\(1.5\times\)
the primary GPU+HBM measurement, compared with \(1.1\)--\(1.25\times\) without
offloading.

\begin{figure}[ht]
  \centering
  \vskip -0.05in
  \includegraphics[width=\linewidth]{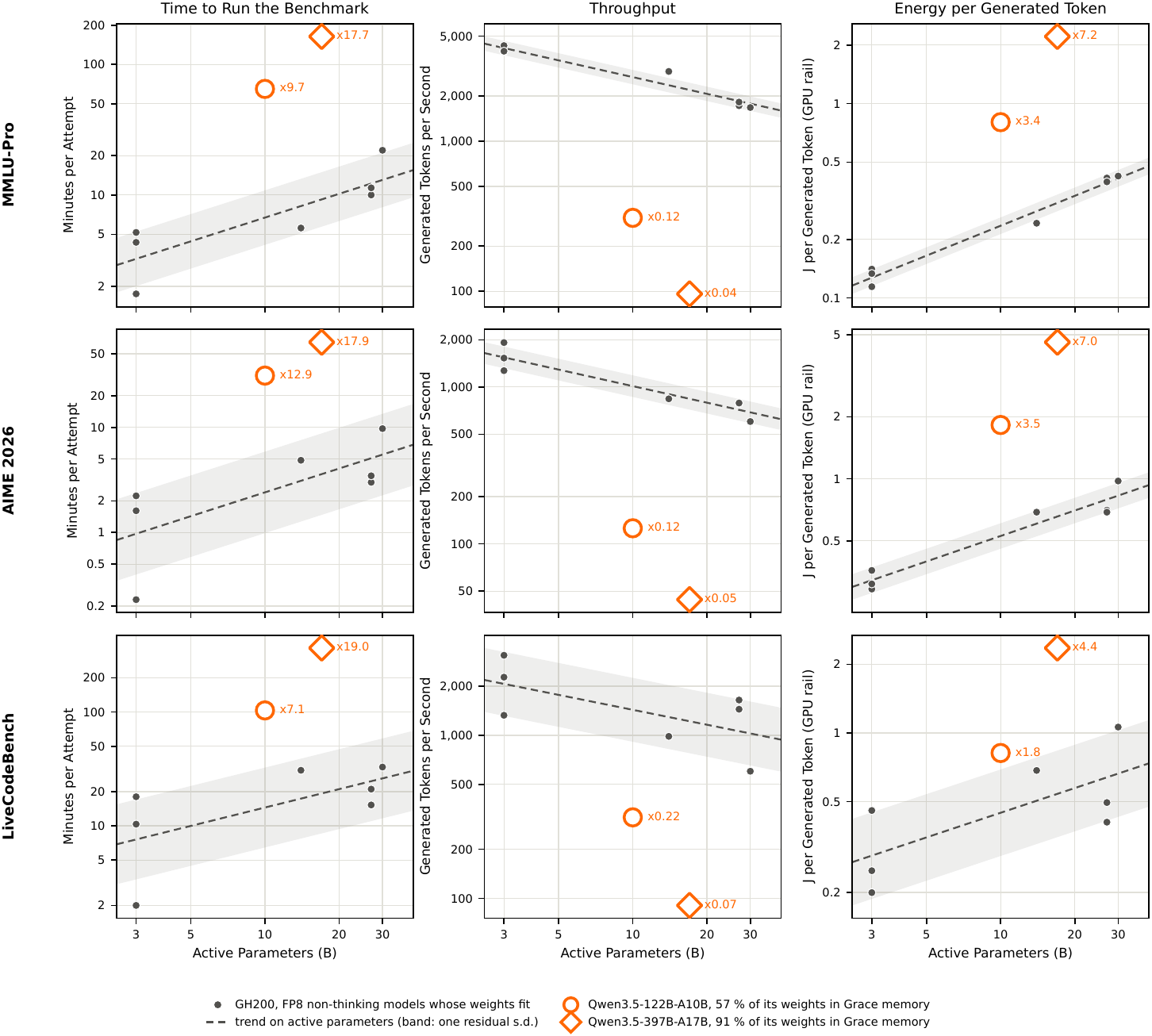}
  \vskip -0.05in
\caption{\textbf{Weights offloaded to Grace memory on the GH200.}
Minutes per attempt (\emph{left}), throughput (\emph{middle}), and energy per
generated token from the GPU+HBM measurement (\emph{right}) against active
parameter count for MMLU-Pro, AIME~2026, and LiveCodeBench. Small points show
FP8 non-thinking configurations that fit entirely in HBM3; dashed lines and
bands show their log-scale least-squares trends and $\pm1$ residual standard
deviation. Open markers show Qwen3.5-122B-A10B (57\% offloaded) and
Qwen3.5-397B-A17B (91\%), labelled by their ratio to the fitted trend.}
  \vskip -0.05in
  \label{fig:offload}
\end{figure}

\clearpage
\section{Energy Consumed by This Study}\label{app:study_energy}

\begin{figure}[ht]
  \centering
  \vskip -0.05in
  \includegraphics[width=\linewidth]{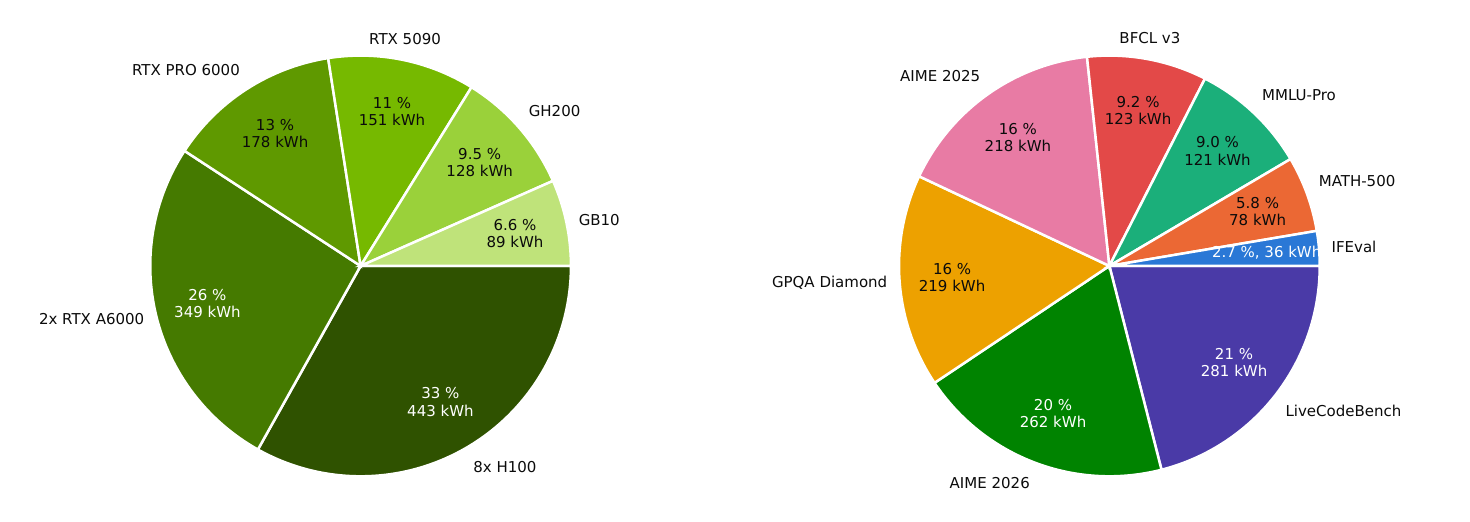}
  \vskip -0.05in
\caption{\textbf{Primary accelerator energy consumed by the completed evaluation
campaign.}
Energy is broken down by platform (\emph{left})
and benchmark (\emph{right}), totalling approximately 1.34\,MWh. The accounting
includes all metered benchmark attempts, including side experiments and attempts
later replaced. Model loading, idle periods, host power, and other infrastructure
overheads are excluded.}
  \vskip -0.05in
  \label{fig:campaign_energy}
\end{figure}

Fig.~\ref{fig:campaign_energy} accounts for the primary accelerator energy
consumed by the completed evaluation campaign. Across all platforms and
benchmarks, the experiments consumed approximately 1.34\,MWh. This total
includes all metered benchmark attempts, including side experiments and attempts
whose results were later replaced or discarded.

The 8$\times$H100 node accounts for the largest share, at 443\,kWh (33\%),
followed by the 2$\times$RTX~A6000 platform at 349\,kWh (26\%). The RTX
PRO~6000, RTX~5090, GH200, and GB10 account for 178, 151, 128, and 89\,kWh,
respectively. By benchmark, LiveCodeBench and AIME~2026 each account for about
one fifth of the total, followed by AIME~2025 and GPQA Diamond at about 16\%
each; IFEval accounts for less than 3\%.

These values use the primary accelerator-level measurement boundary defined in
Appendix~\ref{app:energy}. They therefore exclude power-supply
losses, cooling, host power on the discrete-GPU systems, model-loading energy,
and idle periods between runs. At \$0.20/kWh, the measured accelerator energy
corresponds to approximately \$268 of electricity.

\end{document}